\pdfoutput=1

\documentclass[11pt]{article}

\usepackage{xurl} \usepackage[breaklinks=true]{hyperref}
\usepackage[final]{acl}
\usepackage{amsmath}

\usepackage{times}
\usepackage{latexsym}
\usepackage{booktabs}
\usepackage{amsfonts}

\usepackage[T1]{fontenc}

\usepackage[utf8]{inputenc}

\usepackage{microtype}

\usepackage{inconsolata}

\usepackage{graphicx}
\usepackage{placeins}
\usepackage{pifont}
\usepackage{makecell}
\usepackage{adjustbox}
\usepackage{arydshln}
\usepackage{longtable}

\definecolor{thinkgray}{RGB}{245,245,245}

\usepackage{amsmath}
\usepackage{tcolorbox}
\usepackage{enumitem}
\usepackage{xcolor}
\usepackage{fancyvrb}
\usepackage{float} 

\newtcolorbox{thinkbox}[1][]{%
  colback=thinkgray,
  colframe=black!50,
  sharp corners,
  boxrule=0.5pt,
  #1
}

\newtcolorbox{thinkboxtitle}[2][]{%
  colback=thinkgray,
  colframe=black!50,
  title={#2},
  fonttitle=\normalsize, 
  #1
}

\title{Learning a Continue-Thinking Token for Enhanced Test-Time Scaling}

\author{Liran Ringel\thanks{Equal contribution.}$^{1}$ ~~~~~ Elad Tolochinsky\footnotemark[1]$^{1}$ ~~~~~ Yaniv Romano$^{1,2}$\\
$^1$Department of Computer Science, Technion -- Israel Institute of Technology\\
$^2$Department of Electrical and Computer Engineering, Technion -- Israel Institute of Technology\\
\small{\texttt{liranringel@cs.technion.ac.il},\hspace{0.05cm} \texttt{elad.t@cs.technion.ac.il},\hspace{0.05cm} \texttt{yromano@technion.ac.il}}
}

\begin{document}
\maketitle

\begin{abstract}
Test-time scaling has emerged as an effective approach for improving language model performance by utilizing additional compute at inference time. Recent studies have shown that overriding end-of-thinking tokens (e.g., replacing \texttt{</think>} with ``Wait'') can extend reasoning steps and improve accuracy. In this work, we explore whether a dedicated \emph{continue-thinking} token can be \emph{learned} to trigger extended reasoning. We augment distilled versions of \texttt{DeepSeek-R1} with a single learned \texttt{<|continue-thinking|>} token, training only its embedding via reinforcement learning while keeping the model weights frozen. Our experiments show that this learned token achieves improved accuracy on standard math benchmarks compared to both the baseline model and a test-time scaling approach that uses a fixed token (e.g., ``Wait'') for budget forcing. In particular, we observe that in cases where the fixed-token approach enhances the base model's accuracy, our method achieves a markedly greater improvement. For example, on the GSM8K benchmark, the fixed-token approach yields a 1.3\% absolute improvement in accuracy, whereas our learned-token method achieves a 4.2\% improvement over the base model that does not use budget forcing.
\end{abstract}

\begin{figure*}
    \centering
    \includegraphics[width=0.8\linewidth]{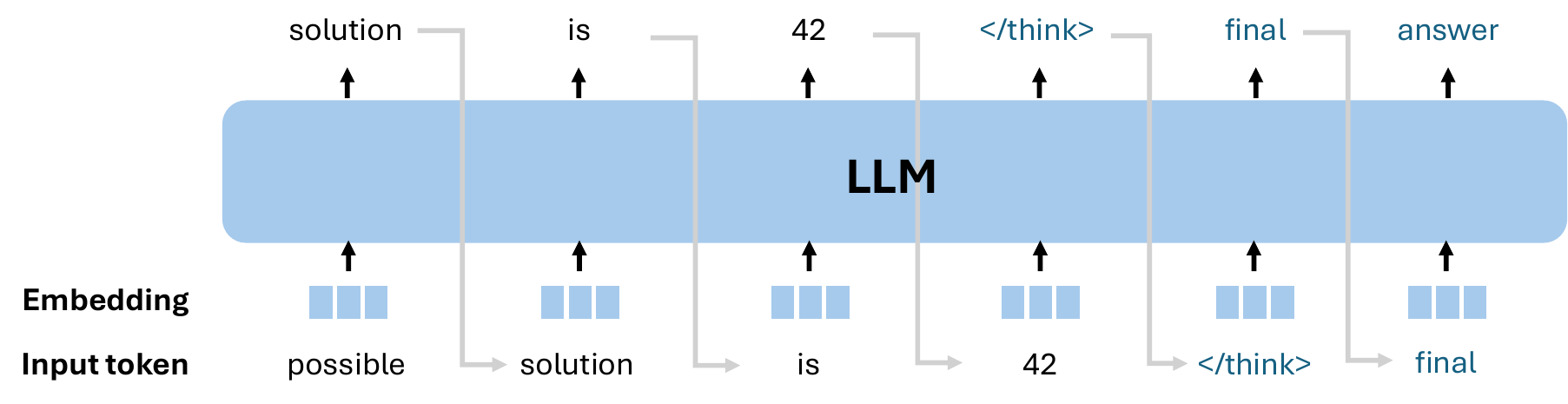}
    
    \vspace{0.5cm}
    
    \includegraphics[width=0.8\linewidth]{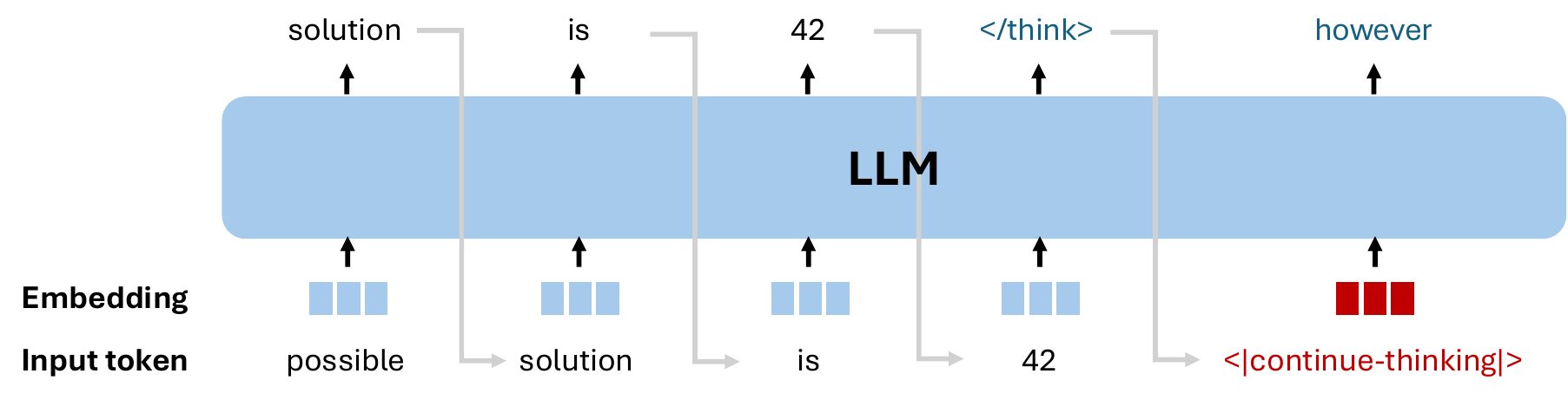}
    \includegraphics[width=0.8\linewidth]{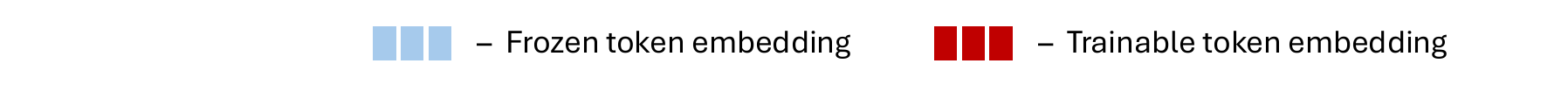}\caption{Text generation with budget forcing: Whenever the model outputs a \texttt{</think>} token, we replace it with a \texttt{<|continue-thinking|>} token and feed that to the model.}
    \label{fig:overview}
\end{figure*}
\section{Introduction}
Language models have demonstrated impressive reasoning abilities through test-time compute scaling~\cite{snell2024scaling, welleck2024decoding, o1, claude_37_thinking}. Two dominant paradigms have emerged for this process \cite{claude_37_thinking}: \textit{parallel} and \textit{sequential}.
The \textit{parallel} approach involves generating multiple samples and selecting the best response based on a majority vote \cite{lewkowycz2022solving} or a model-provided score \cite{claude_37_thinking}. In contrast, the \textit{sequential} approach---central to this work and popularized by OpenAI's o1 model \cite{o1}---generates a single sample and encourages the model to revisit, backtrack, validate, and refine its reasoning before producing a final answer, typically resulting in a long chain-of-thought output \cite{wei2022chain}. Models that follow this paradigm are commonly referred to as \textit{reasoning models}, and they are often trained using reinforcement learning with verifiable rewards~\cite{lambert2024tulu,guo2025deepseek,su2025expanding,wang2025reinforcement}.

A key property of reasoning models is their ability to decide when to stop thinking, typically by generating an explicit end-of-thinking token (e.g., \texttt{</think>}). However, since this decision is made by the model, users have no direct control over the amount of reasoning performed. Recently, budget forcing was introduced in \textit{s1: simple test-time scaling} \cite{muennighoff2025s1}, a sequential test-time scaling approach that provides direct control over the model’s computation time. By replacing end-of-thinking tokens with ``Wait'' tokens during generation, the authors showed that longer chain-of-thought reasoning could be achieved, resulting in improved accuracy. Conversely, early termination can be enforced by appending a \texttt{</think>} token once the model reaches its compute budget, prompting it to generate the final answer. Budget forcing was quickly shown to perform well in various settings \cite{aggarwal2025l1, huang2025m1}; see \autoref{sec:related_work} for a detailed discussion.

In this paper, we introduce a systematic approach for learning a special \texttt{<|continue-thinking|>} token as an alternative to the ``Wait'' or related fixed tokens such as ``Alternatively'' or ``Hmm'' suggested by \citet{muennighoff2025s1}. Our primary goal is to rigorously investigate whether the simple practice of using a fixed ``Wait'' token for budget forcing can be improved by learning a dedicated token embedding.
In line with this goal, we deliberately constrain ourselves to the same budget forcing algorithm.

As illustrated in \autoref{fig:overview}, we introduce a special, learned \texttt{<|continue-thinking|>} token into the model's vocabulary. During generation, we modify the model's output generation process to replace any occurrence of the \texttt{</think>} token with our \texttt{<|continue-thinking|>} token, as long as the token budget is not exhausted and the maximum number of forced continuations has not been reached. During training, we optimize only the embedding vector of the \texttt{<|continue-thinking|>} token while keeping all other model parameters frozen. 

We train this new token using reinforcement learning (RL), specifically the group relative policy optimization (GRPO) procedure~\cite{shao2024deepseekmath}. While supervised fine-tuning is a possible alternative, designing effective supervised reasoning demonstrations is difficult. In contrast, RL allows the training process to explore novel continuations aimed at improving task accuracy, making it a more natural fit for our setting. Notably, our method leverages a frozen model backbone, optimizing only a single parameter vector of the size of the model’s hidden dimension.
As a result, the optimizer's additional memory overhead is limited to just this vector, rather than all model parameters. This allows us to allocate more GPU memory for longer context windows during RL training.

We apply our proposed method to investigate the influence of the learned \texttt{<|continue-thinking|>} token on the reasoning process. Our results show that learning this token can significantly enhance model performance, yielding greater gains than the ``Wait'' or related tokens used in s1's budget-forcing method. Notably, we find that whenever budget-forcing with ``Wait'' provides an improvement over the baseline, our learned token achieves even greater gains---up to a 320\% increase in relative accuracy improvements, and as much as a 4\% absolute improvement in overall accuracy. Conversely, in cases where budget-forcing with a fixed token (such as ``Wait'') does not improve performance compared to the baseline, the learned token similarly does not offer a statistically significant benefit. Results are reported in \autoref{tab:results}.

\begin{table*}[ht]
\begin{small}
\begin{tabular}{l l l l l l l} 
 \toprule
 Dataset  & Baseline  & Alternatively & Hmm & Wait & Critique & Learned \\ 
 & (w/o BF)\\
  \midrule
  \texttt{DS-R1-Qwen-1.5B} \\
 \midrule
 $C_{\text{max}} \!= \!3$ \\
 AIME24 & 23.85 $\pm$ 0.66  & 23.49 $\pm$ 0.69  & 24.48 $\pm$ 0.63  & 23.91 $\pm$ 0.70  & 23.12  $\pm$ 0.71  & 24.06 $\pm$ 0.68  \\
AIME25 & 23.33 $\pm$ 0.52  & 23.80 $\pm$ 0.56  & 23.85 $\pm$ 0.53  & 23.49 $\pm$ 0.49  & 22.45  $\pm$ 0.49  & 22.40 $\pm$ 0.44  \\
GSM8K & 78.08 $\pm$ 0.36  & 78.83 $\pm$ 0.34  & 79.47 $\pm$ 0.34  & 80.14 $\pm$ 0.24  & 76.10 $\pm$ 0.59 & \textbf{83.68}  $\pm$ 0.38  \\
MATH500 & 79.88 $\pm$ 0.33  & 80.33 $\pm$ 0.35  & 80.84 $\pm$ 0.39  & 81.22 $\pm$ 0.30 & 79.49  $\pm$ 0.32  & \textbf{82.51}  $\pm$ 0.34  \\
BBH & 45.71 $\pm$ 0.60  & 46.75 $\pm$ 0.43  & 47.06 $\pm$ 0.60  & 47.58 $\pm$ 0.52 & 46.00  $\pm$ 0.72  & 47.70  $\pm$ 0.62  \\
  \midrule

\texttt{DS-R1-Qwen-7B} \\
 \midrule
 $C_{\text{max}}  \!= \! 3$ \\
AIME24 & 46.09 $\pm$ 0.80  & 46.46 $\pm$ 0.73  & 47.24 $\pm$ 0.71  & 47.19 $\pm$ 0.77  & 46.41 $\pm$ 0.75  & 46.82 $\pm$ 0.76  \\
AIME25 & 36.41 $\pm$ 0.66  & 36.51 $\pm$ 0.57  & 36.51 $\pm$ 0.61  & 36.77 $\pm$ 0.63  & 37.29 $\pm$ 0.71  & 36.09 $\pm$ 0.58  \\
GSM8K & 88.81 $\pm$ 0.30  & 89.27 $\pm$ 0.27  & 91.23 $\pm$ 0.26  & 91.74 $\pm$ 0.34  & 88.89 $\pm$ 0.39  & \textbf{94.91}  $\pm$ 0.18  \\
MATH500 & 89.39 $\pm$ 0.31  & 89.84 $\pm$ 0.27  & 91.74 $\pm$ 0.19  & 92.17 $\pm$ 0.17  & 89.06 $\pm$ 0.27  & \textbf{92.89}  $\pm$ 0.20  \\
BBH & 71.05 $\pm$ 0.75  & 72.53 $\pm$ 0.72  & 72.91 $\pm$ 0.63  & 73.85 $\pm$ 0.51 & 71.01  $\pm$ 0.76  & 74.75  $\pm$ 0.60  \\
\midrule

\texttt{DS-R1-Llama-8B} \\
 \midrule
 $C_{\text{max}}  \!= \! 3$ \\
AIME24	& 35.68 $\pm$ 0.78	& 36.93 $\pm$ 0.75	& 38.23 $\pm$ 0.73	& 38.39 $\pm$ 0.79	& 35.73 $\pm$ 0.71	&\textbf{ 41.30} $\pm$ 0.91 \\
AIME25	& 29.32 $\pm$ 0.61	& 29.64 $\pm$ 0.57	& 30.10 $\pm$ 0.61	& 30.26 $\pm$ 0.64	& 29.90 $\pm$ 0.58	& 30.26 $\pm$ 0.65 \\
GSM8K	& 78.43 $\pm$ 0.40	& 79.82 $\pm$ 0.41	& 85.22 $\pm$ 0.22	& 84.90 $\pm$ 0.43	& 77.79 $\pm$ 0.65	& \textbf{93.78} $\pm$ 0.10 \\
MATH500	& 79.56 $\pm$ 0.31	& 81.86 $\pm$ 0.39	& 86.21 $\pm$ 0.32	& 86.01 $\pm$ 0.32	& 80.27 $\pm$ 0.40	& \textbf{90.04} $\pm$ 0.25 \\
BBH & 79.73 $\pm$ 0.37  & 81.32 $\pm$ 0.30  & 81.79 $\pm$ 0.46  & 81.97 $\pm$ 0.32 & 80.26  $\pm$ 0.42  & \textbf{83.69}  $\pm$ 0.37  \\
 \bottomrule
\end{tabular}
\end{small}
\caption{Accuracy (pass@1) results for $B_{\text{max}} = 9216$  and different numbers of forced thinking continuations $C_{\text{max}}$. Results are obtained via regex-based evaluation and an LLM evaluator if the model fails to generate an answer in the correct format. See \autoref{tab:results-full-1.5b}, \autoref{tab:results-full-7b}, and \autoref{tab:results-full-llama-8b} for the full results.}
\label{tab:results}
\end{table*}

To summarize, our contributions have the following key features:
\begin{itemize}
\item We introduce the concept of learning a specialized \texttt{<|continue-thinking|>} token as an effective mechanism for budget-controlled compute scaling.
\item Our experiments demonstrate that, in cases where budget-forcing with a fixed token such as ``Wait'' improves accuracy, learning the token yields even greater gains, while requiring only a single additional token embedding.
\item We use an external LLM to compare outputs with ground truth, mitigating the limitations of standard math benchmark evaluations that rely on rigid answer formats, assuming the final answer is in \verb|\boxed{}|. Our experiments show that this rigid-format evaluation approach can misrepresent reasoning ability.
\end{itemize}

The code used for training and evaluation is available at \url{https://github.com/liranringel/learning-continue-thinking-token}.

\section{Method}

\subsection{Single Token Optimization}

Let $\pi(x)$ be a pretrained language model that takes a prompt $x$ as input and generates an output. Key to our method is the introduction of a special token, \texttt{<|continue-thinking|>}, which we add to the model's vocabulary to promote longer reasoning traces at test time. We denote the embedding vector of this token by $\theta_T \in \mathbb{R}^{d}$, where $d$ is the embedding dimension of the model.

We refer to the adapted model with the new token as $\pi_{\theta_T}$. All model parameters are frozen except for $\theta_T$, so the embedding of the \texttt{<|continue-thinking|>} token is the only parameter updated during training.

Our objective is to maximize the reasoning performance of $\pi_{\theta_T}$ by optimizing the embedding vector $\theta_T$. Formally, we aim to optimize the following:
\begin{equation}
    \label{eq:optimization_problem}
     \theta_T^{*} := \arg\max_{\theta_T} \mathbb{E}_{x \sim Q, o \sim \text{BF}(\pi_{\theta_T}, x)} [R(x, o)],
\end{equation}
where $x$ is a question sampled from a distribution $Q$, and $o$ is the response generated by running the budget forcing algorithm on $x$ using the model $\pi_{\theta_T}$. Note that $o$ is not a standard sample from $\pi_{\theta_T}(y \mid x)$; rather, its distribution is induced by the budget forcing algorithm $\text{BF}(\pi_{\theta_T}, x)$, which we present below. The function $R(x, o)$ represents the reward associated with the generated output. For example, for mathematical questions, $R(x, o)$ could be an indicator function that equals $1$ if $o$ contains the correct answer to question $x$, and $0$ otherwise.

In more detail, the budget forcing algorithm $\text{BF}(\pi_{\theta_T}, x)$ modifies the generation process by enforcing additional reasoning steps before producing a final answer. During generation, whenever the model outputs an end-of-thinking \texttt{</think>} token, it is replaced with the learned \texttt{<|continue-thinking|>} token, as a way to force longer reasoning traces. This process repeats until one of the following conditions is met:
\begin{enumerate}    
    \item The number of forced thinking continuations reaches the preset maximum $C_{\text{max}}$.
    \item The total number of generated tokens reaches the budget limit $B_{\text{max}}$, in which case the reasoning process is immediately terminated and a \texttt{</think>} token is inserted. This ensures that the model does not generate beyond the allowed compute budget.
\end{enumerate}

During training, we set $C_{\text{max}} = 1$, meaning that only a single forced continuation is allowed per input. At test time, however, we also evaluate the model with $C_{\text{max}} > 1$ to assess its ability to generalize to multiple forced continuations.

\subsection{Training}
Throughout this paper, we apply our method to distilled versions of DeepSeek-R1, specifically its Qwen 1.5B, Qwen 7B, and Llama 8B variants~\cite{dubey2024llama,yang2024qwen2,guo2025deepseek}, training only the embedding of a newly introduced token, while keeping all other parameters frozen. We set the number of forced continuations to $C_{\text{max}} = 1$ and 
the budget limit to $B_{\text{max}} = 8192$ for the 1.5B model and $B_{\text{max}} = 10240$ for larger models. We initialize the new token’s embedding using that of the word ``Wait.''
The reward function used during training is the sum of two binary components: (i) a format reward that verifies whether the answer is in the expected format (specifically, wrapped in \verb|\boxed{}|), and (ii) a correctness reward, which checks whether the generated answer matches the ground truth. We implemented our training code on top of Open-R1~\cite{openr1} and TRL~\cite{vonwerra2022trl}. For GRPO, we set $G=16$ generations per example and used a batch size of 16. We performed 64 gradient accumulation steps, resulting in an effective batch size of 1,024 generations (i.e., 64 examples with 16 generations each) per optimizer update. The training dataset is \texttt{DeepScaleR-Preview-Dataset}~\cite{deepscaler2025}, which is a collection of 40,000 math questions compiled from various datasets. We used 8 NVIDIA A100 GPUs with 80GB memory for training the 1.5B model and 8 NVIDIA H200 GPUs for training the larger models.  The total training time for each model was about 5 days. See \autoref{tab:parameters} for a full list of parameters and system prompts used during training.

\section{Related Work}
\label{sec:related_work}
Budget forcing, introduced by \citet{muennighoff2025s1}, is a simple and effective method for scaling compute at test time. L1~\cite{aggarwal2025l1} extends this idea using reinforcement learning to train models that satisfy user-specified reasoning lengths, enabling flexible cost-performance trade-offs. The m1 method by~\citet{huang2025m1} further explores the application of budget-forcing in medical QA. \citet{jin2025well} introduce a variant of budget forcing for greedy decoding, comparing multi-word phrases instead of single tokens like ``Wait.'' Collectively, these works highlight budget forcing as a promising direction for compute budget control.

\begin{table*}[ht]
\begin{center}
\begin{small}
\begin{adjustbox}{width=\linewidth}
\begin{tabular}{l l l l l l l} 
 \toprule
 Dataset  & Baseline  & Alternatively & Hmm & Wait & Critique & Learned \\ 
 & (w/o BF)\\
  \midrule
  \texttt{DS-R1-Qwen-1.5B} \\
  \midrule
 $C_{\text{max}} = 2$  \\
 AIME25 & 23.18 $\pm$ 0.50 (86) & 23.02 $\pm$ 0.49 (85) & 23.23 $\pm$ 0.51 (86) & 23.23 $\pm$ 0.50 (87)& 22.14  $\pm$ 0.49 (85) & 22.80 $\pm$ 0.47 (86) \\
 GSM8K & 66.03 $\pm$ 0.40 (84) & 64.82 $\pm$ 0.25 (82) & 64.09 $\pm$ 0.38 (80) & 64.43 $\pm$ 0.48 (79) & 69.99  $\pm$ 0.44 (88) & \textbf{78.27} $\pm$ 0.58 (94) \\
 MATH500  & 77.56 $\pm$ 0.32 (95)& 77.97 $\pm$ 0.35 (95)& 78.88 $\pm$ 0.32 (95)& 78.83 $\pm$ 0.38 (95)& 77.56  $\pm$ 0.28 (95) & \textbf{81.37} $\pm$ 0.38 (97)\\
 \hline
 $C_{\text{max}} = 3$ & & & & & \\
AIME25 & 23.18 $\pm$ 0.50 (85) & 23.44 $\pm$ 0.56 (87) & 23.49 $\pm$ 0.49 (86) & 23.12 $\pm$ 0.46 (86) & 22.14  $\pm$ 0.46 (86) & 22.40 $\pm$ 0.44 (86) \\
GSM8K & 66.03 $\pm$ 0.40 (83) & 63.79 $\pm$ 0.57 (80) & 64.27 $\pm$ 0.28 (79) & 62.94 $\pm$ 0.50 (77) & 69.35  $\pm$ 0.56 (87) & \textbf{80.87}  $\pm$ 0.42 (96) \\
MATH500 & 77.56 $\pm$ 0.32 (94) & 78.16 $\pm$ 0.39 (95) & 78.60 $\pm$ 0.40 (94) & 78.75 $\pm$ 0.32 (94) & 78.08  $\pm$ 0.28 (96) & \textbf{81.75}  $\pm$ 0.34 (96) \\
 \midrule
   \texttt{DS-R1-Qwen-7B} \\
  \midrule
 $C_{\text{max}} = 2$  \\
AIME24 & 44.74 $\pm$ 0.77 (71) & 45.47 $\pm$ 0.78 (71) & 45.94 $\pm$ 0.78 (72) & 45.62 $\pm$ 0.79 (72) & 45.42 $\pm$ 0.77 (71) & 45.52 $\pm$ 0.80 (72) \\
AIME25 & 35.47 $\pm$ 0.62 (67) & 35.68 $\pm$ 0.59 (68) & 35.00 $\pm$ 0.62 (66) & 35.62 $\pm$ 0.66 (67) & 36.82 $\pm$ 0.67 (67) & 34.90 $\pm$ 0.60 (68) \\
GSM8K & 88.81 $\pm$ 0.30 (99) & 88.82 $\pm$ 0.39 (99) & 90.58 $\pm$ 0.29 (99) & 90.85 $\pm$ 0.39 (99) & 89.12 $\pm$ 0.14 (99) & \textbf{94.64}  $\pm$ 0.19 (99) \\
MATH500 & 88.94 $\pm$ 0.30 (97) & 88.90 $\pm$ 0.24 (97) & 91.08 $\pm$ 0.19 (97) & 91.24 $\pm$ 0.20 (97) & 88.74 $\pm$ 0.26 (97) & \textbf{92.65}  $\pm$ 0.18 (97) \\
 \hline
 $C_{\text{max}} = 3$  \\
AIME24 & 44.74 $\pm$ 0.77 (71) & 45.52 $\pm$ 0.72 (72) & 46.15 $\pm$ 0.72 (71) & 46.30 $\pm$ 0.74 (73) & 45.21 $\pm$ 0.80 (73) & 45.94 $\pm$ 0.79 (72) \\
AIME25 & 35.47 $\pm$ 0.62 (67) & 35.47 $\pm$ 0.57 (66) & 35.52 $\pm$ 0.62 (67) & 35.73 $\pm$ 0.64 (66) & 36.41 $\pm$ 0.68 (67) & 35.16 $\pm$ 0.58 (66) \\
GSM8K & 88.81 $\pm$ 0.30 (99) & 89.15 $\pm$ 0.25 (99) & 90.45 $\pm$ 0.24 (98) & 91.51 $\pm$ 0.35 (99) & 88.74 $\pm$ 0.38 (99) & \textbf{94.84}  $\pm$ 0.18 (99) \\
MATH500 & 88.94 $\pm$ 0.30 (97) & 89.42 $\pm$ 0.27 (97) & 91.20 $\pm$ 0.16 (97) & 91.78 $\pm$ 0.18 (97) & 88.70 $\pm$ 0.26 (97) & \textbf{92.62}  $\pm$ 0.20 (97) \\
 \bottomrule
\end{tabular}
\end{adjustbox}
\end{small}
\end{center}
\caption{Accuracy (pass@1) results for $B_{\text{max}} = 9216$ and different numbers of forced thinking continuations $C_{\text{max}}$. Results are obtained via a regex-based evaluation only. The percentage of final answers enclosed in \texttt{\textbackslash boxed\{\}} is shown in parentheses. See \autoref{tab:results_no_llm_full-1.5b} and \autoref{tab:results_no_llm_full-7b} for the full results.}
\label{tab:results_no_llm}
\end{table*}

The concept of incorporating additional tokens has been explored in several prior studies. The works reported in \cite{goyal2024think} and \cite{wang2023guiding} introduce learnable tokens into reasoning traces to improve the model's accuracy. In \cite{goyal2024think,pfau2024let}, additional tokens are inserted at random positions during training and appended to the prompt during inference, allowing the model to artificially increase the number of activations at test time. \citet{wang2023guiding} included a `planning token' at the start of each reasoning step.
Unlike our approach, these methods rely on supervised fine-tuning to learn token representations. In contrast, our method utilizes RL to optimize the new token embedding.

Finally, various works on scaling test time compute motivate our choice of token learning using RL: (1) Recent empirical findings show that RL-based fine-tuning leads to better generalization \cite{chu2025sft}; (2) theoretical analysis shows that RL enjoys higher expected cumulative reward \cite{setlur2025scaling}, and (3) learning to use additional tokens, which is related to our approach of designing a \texttt{<|continue-thinking|>} token, has been observed to be a hard learning problem when using supervised learning \cite{pfau2024let}.

\begin{figure*}[ht]
    \centering
    \includegraphics[width=\linewidth]{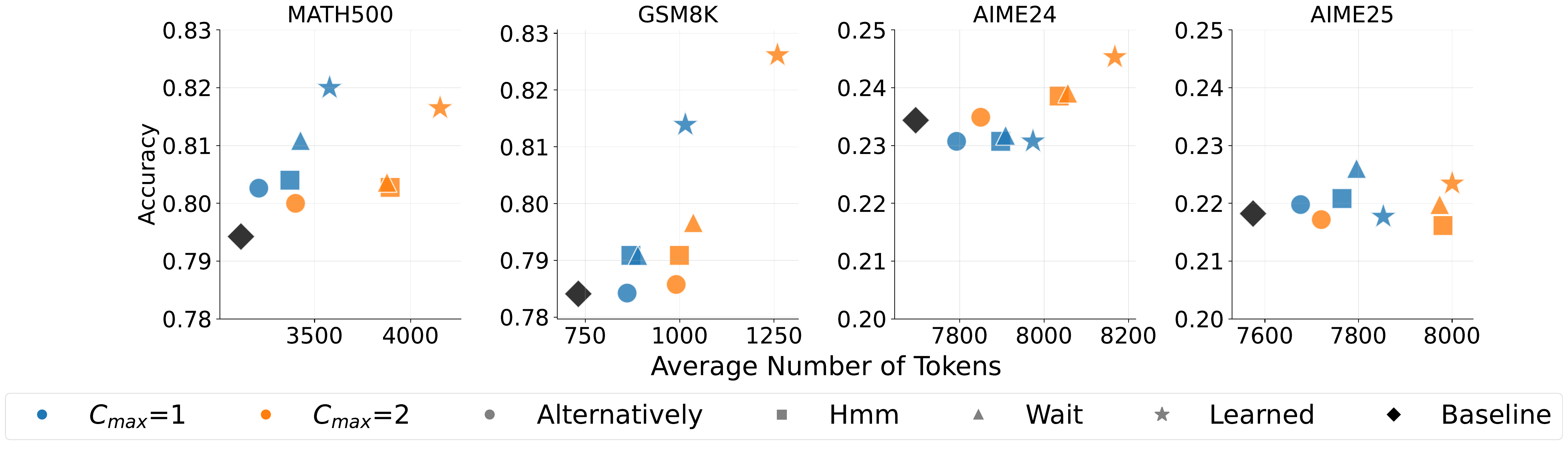}
 \caption{Accuracy of different methods as a function of the average number of tokens generated by each method. Results for all datasets are obtained using \texttt{DeepSeek-R1-Distill-Qwen-1.5B} and $B_{\text{max}} = 8192$. }
    \label{fig:average_tokens}
\end{figure*}

\section{Experiments}

\subsection{Evaluation protocol}
We evaluate our model on three widely adopted mathematical reasoning datasets: GSM8K-Platinum~\cite{cobbe2021training, vendrow2025largelanguagemodelbenchmarks}, a revised version of the original GSM8K dataset containing 1209 grade-school level math problems; MATH500 \cite{lightman2023let, muennighoff2025s1}, a 500-question subset of the MATH \cite{hendrycks2021measuring} dataset; and 
AIME24~\cite{muennighoff2025s1} and AIME25 datasets, which contain 30 math problems from the 2024 and 2025 American Invitational Mathematics Examination, a national-level mathematics competition in the United States. We implemented the evaluation pipeline using a modified version of the LM-Evaluation-Harness library~\cite{eval-harness}.

To further assess the generalization of our approach beyond mathematical reasoning, we evaluated all models on a subset of 16 tasks from BIG-Bench Hard (BBH)~\cite{suzgun2022challenging}, a benchmark comprising challenging problems that probe symbolic, logical, and commonsense reasoning abilities. To ensure reliable automatic evaluation, we restricted BBH to questions with multiple-choice answers. Importantly, all models were trained exclusively on math questions, so performance on BBH provides a measure of cross-domain generalization.

Our method is compared against three fixed tokens that have previously demonstrated strong performance \cite{muennighoff2025s1}, a baseline configuration that does not employ budget-forcing, and an explicit guiding prompt: "Critique your previous step and try again".

Due to the relatively small size of some of the datasets, we generate multiple responses per question to compute standard error as a way to enhance the statistical reliability of the reported results. Specifically, we found that generating the following number of completions produces error bars that allow us to distinguish between statistically significant and insignificant results: 16 samples per question for MATH500, 64 for AIME, 6 samples for GSM8K, and 10 samples for BBH. We report (pass@1) accuracy along with the standard error for each setting. Accuracy is computed using a regex-based evaluation script that extracts the final answer from the model’s output and checks for an exact match with the ground truth. The evaluation returns `True' if the extracted answer is correct, and `False' otherwise---including cases where the regex fails to match a valid answer. In case the regex fails to match, we employ an external LLM to assess if the generated answer was semantically equivalent to the ground truth. This hybrid evaluation strategy was adopted based on our manual inspection, which indicated that regex matching is more reliable than LLM-based comparison when the regex successfully parses the output; see \autoref{sec:llm_verfication} for a detailed discussion. We utilized \texttt{Qwen/Qwen2.5-7B-Instruct}~\cite{yang2024qwen2technical} as our evaluator LLM. See \autoref{tab:parameters} for the instruction prompt we used for the LLM evaluation. 

During inference, we verify the generalization of our method by also using inference configurations not used during training. Concretely, we set the reasoning budgets $B_{\text{max}} = 8192, 9216$ and the maximal number of forced continuations $C_{\text{max}} = 1, 2, 3$. We also provide an additional 1,024 tokens reserved for generating the final answer. For inference, we used the same system prompt as the one used for training; see \autoref{tab:parameters} in the Appendix. Due to the prolonged length of the generated answers, evaluation took approximately 370 GPU hours, using 8 NVIDIA A40 GPUs. 

\begin{figure*}
    \centering
    \includegraphics[width=\linewidth]{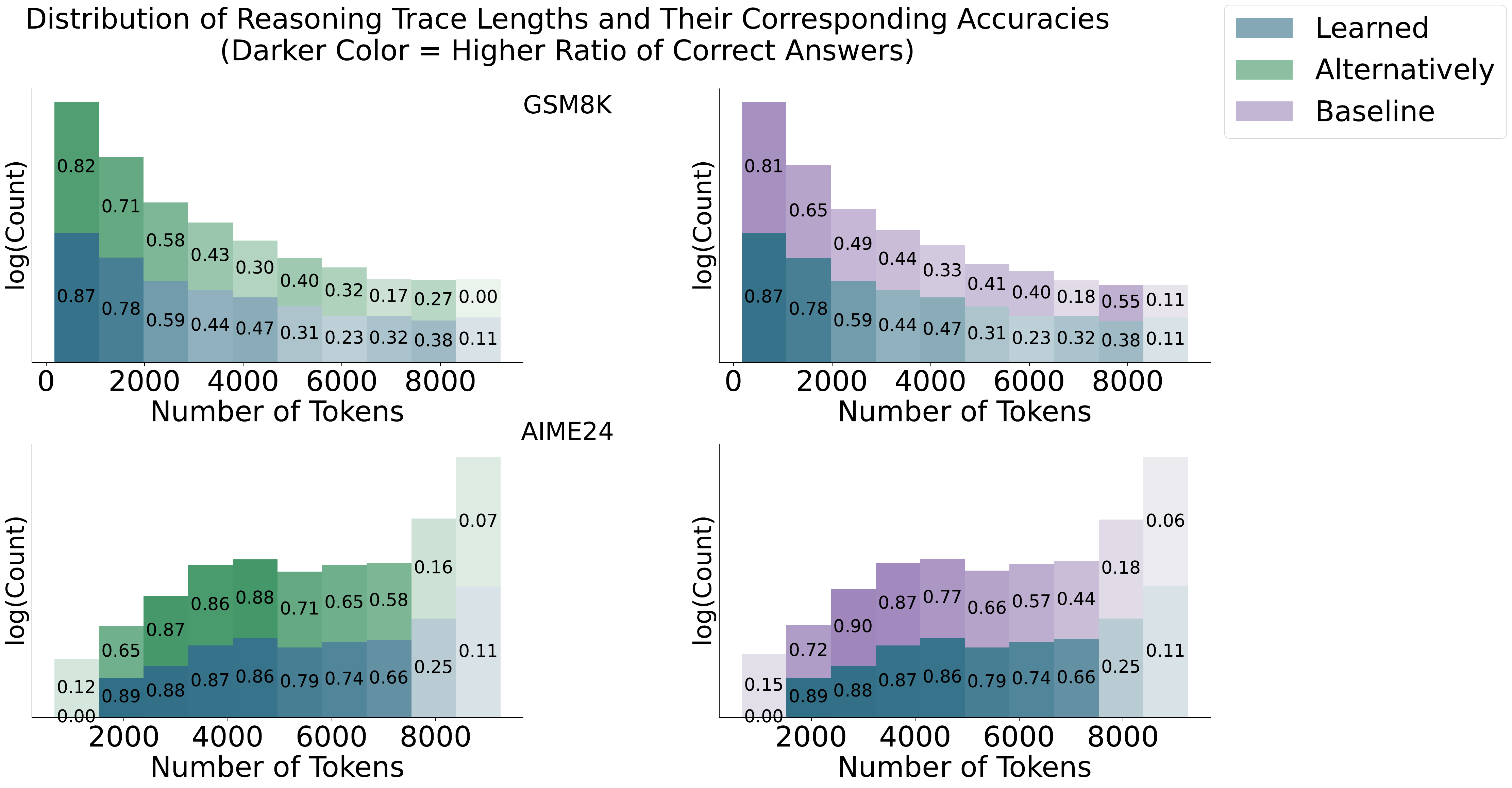}
 \caption{Comparison of generated sequence length distributions and their corresponding accuracies. Stacked bars represent the logarithmic count of answers within each length bin, with darker segments indicating a higher proportion of correct answers (fraction shown within each bin). Top row: GSM8K, Bottom row: AIME24. Left: Learned \texttt{<|continue-thinking|>} token vs. ``Alternatively.'' Right: Learned \texttt{<|continue-thinking|>} token vs. baseline model without budget forcing. Data was obtained using \texttt{DS-R1-Qwen-1.5B} and $C_{\text{max}} \!= \!2, B_{\text{max}} \!=\! 8192$.}
\label{fig:token_distribution}
\end{figure*}

\newcommand{\x}{\textcolor{red}{\ding{55}}}
\newcommand{\tick}{\textcolor{green}{\ding{51}}}

\begin{table*}[t]
\begin{center}
\begin{small}
\begin{tabular}{l l l l l} 
 \toprule
 Probabilities & AIME24 & AIME25 & GSM8K & MATH500 \\  
  \midrule
 $C_{\text{max}} = 2$, $B_{\text{max}} = 8192$ & & & &  \\
 P(Learned \x, Baseline \tick) & 0.01 & 0.01 & 0.03 & 0.02 \\
 P(Learned \tick, Baseline \x) & 0.02 & 0.02 & 0.07 & 0.04 \\
 \hline
 $C_{\text{max}} = 3$, $B_{\text{max}} = 8192$ & & & &  \\
 P(Learned \x, Baseline \tick) & 0.09 & 0.06 & 0.02 & 0.01 \\
 P(Learned \tick, Baseline \x) & 0.09 & 0.06 & 0.07 & 0.04 \\
 \hline
 $C_{\text{max}} = 2$, $B_{\text{max}} = 9216$ & & & &  \\
 P(Learned \x, Baseline \tick) & 0.08 & 0.06 & 0.02 & 0.02 \\
 P(Learned \tick, Baseline \x) & 0.08 & 0.06 & 0.07 & 0.04 \\
 \hline
 $C_{\text{max}} = 3$, $B_{\text{max}} = 9216$ & & & &  \\
 P(Learned \x, Baseline \tick) & 0.09 & 0.06 & 0.02 & 0.02 \\
 P(Learned \tick, Baseline \x) & 0.09 & 0.06 & 0.07 & 0.04 \\
 \bottomrule
\end{tabular}
\end{small}
\end{center}
\caption{Probability of observing a correct answer from the learned token and an incorrect answer from the baseline and vice versa when using \texttt{DeepSeek-R1-Distill-Qwen-1.5B}. See \autoref{tab:probabilities-full} in the Appendix for the full results.}
\label{tab:probabilities}
\end{table*}

\subsection{LLM-Based Verification}
\label{sec:llm_verfication}

It is common practice to use regex-based functions to check the correctness of model outputs when evaluating language models on mathematical benchmarks.
However, this approach can distort the evaluation of model performance, as it conflates output formatting with actual reasoning ability.
In our experiments, evaluating with only regex-based functions suggested that the learned token led to substantial performance gains. However, a more detailed analysis using an evaluator LLM revealed that much of this improvement was due to better adherence to formatting rather than genuine reasoning improvements.
As shown in \autoref{tab:results_no_llm}, the results obtained using only regex-based evaluation are significantly higher than those reported in \autoref{tab:results}, which use LLM-based evaluation.
For example, on GSM8K with $C_{\text{max}} = 2$ and $B_{\text{max}} = 8192$, the improvement of our method over the baseline is more than three times higher when using regex-based evaluation compared to LLM-based evaluation. Similarly, on the MATH500 dataset with $C_{\text{max}} = 1$ and $B_{\text{max}} = 9216$, regex-based evaluation indicates a statistically significant improvement, whereas LLM-based evaluation shows that this improvement is not actually present.
 
To further assess the trustworthiness of our evaluation procedure, we manually inspected answers generated by $\texttt{DS-R1-Qwen-7B}$ along with their corresponding LLM-based evaluation results for 80 randomly selected questions from the MATH500 dataset and found that the LLM evaluation aligned with human assessment in all cases.
 
Nonetheless, we acknowledge that LLM-based verification is itself imperfect. We adopt it as a more reliable alternative to purely regex-based evaluation, improving the robustness and credibility of our reported results.

\subsection{Results}
The evaluation results are depicted in \autoref{tab:results}. Our findings indicate that when budget forcing does not improve upon the baseline, the learned token similarly offers no statistically significant advantage. However, on datasets where budget forcing yields improvements, our learned token demonstrates a substantial performance increase, achieving up to a 320\% relative gain over the best fixed token and a 4\% absolute improvement in accuracy.
Notably, although our model was trained with $C_{\text{max}} = 1$, we observe that increasing $C_{\text{max}}$ to 2 and 3 during inference often leads to further improvements in accuracy. This suggests that the learned \texttt{<|continue-thinking|>} token can generalize to settings with multiple forced continuations, even though the model was not explicitly trained for them.

To gain deeper insights into the behavior of the different methods, we analyzed the generated token distributions. \autoref{fig:average_tokens} depicts the accuracy of each method as a function of the average number of tokens generated per dataset. We observed that using the learned token consistently resulted in longer reasoning traces, suggesting that the performance improvement can be attributed to this increased reasoning length. Furthermore, \autoref{fig:average_tokens} shows that for the AIME datasets, all methods generated a high average number of tokens. This can likely be attributed to the significant difficulty of the AIME problems for our model, which might also explain why our method did not yield substantial improvements on these datasets.

Indeed, \citet{wu2025s} has also noted that budget forcing provides inconsistent or minimal benefits on AIME benchmarks. Since AIME consists of only 30 challenging questions, each additional correct answer significantly increases the overall score (approximately +3.3\% per question). Therefore, the moderate improvements provided by budget forcing may be insufficient to consistently reach the threshold required to solve the next level of question difficulty, which might explain the limited effectiveness observed both here and in prior studies.

A comparison of the generated answer length distributions for the learned token, the baseline, and the ``Alternatively'' token is shown in \autoref{fig:token_distribution}. The accuracy improvements observed with the learned token are consistent across different generated lengths, not just on average. This suggests that the enhanced performance is attributable to a genuine improvement in the model's reasoning capabilities, beyond merely generating longer responses.
In \autoref{tab:probabilities}, we show the probabilities that a correct answer under the baseline will be incorrect with the learned token. We can see that, for both GSM8K and MATH500, it is much more likely that the learned token leads to improvement. For the AIME dataset, we see that the learned token and the baseline are comparable, which is expected, since both methods have similar accuracy.

\begin{figure}[h]
    \centering
    \includegraphics[width=\linewidth]{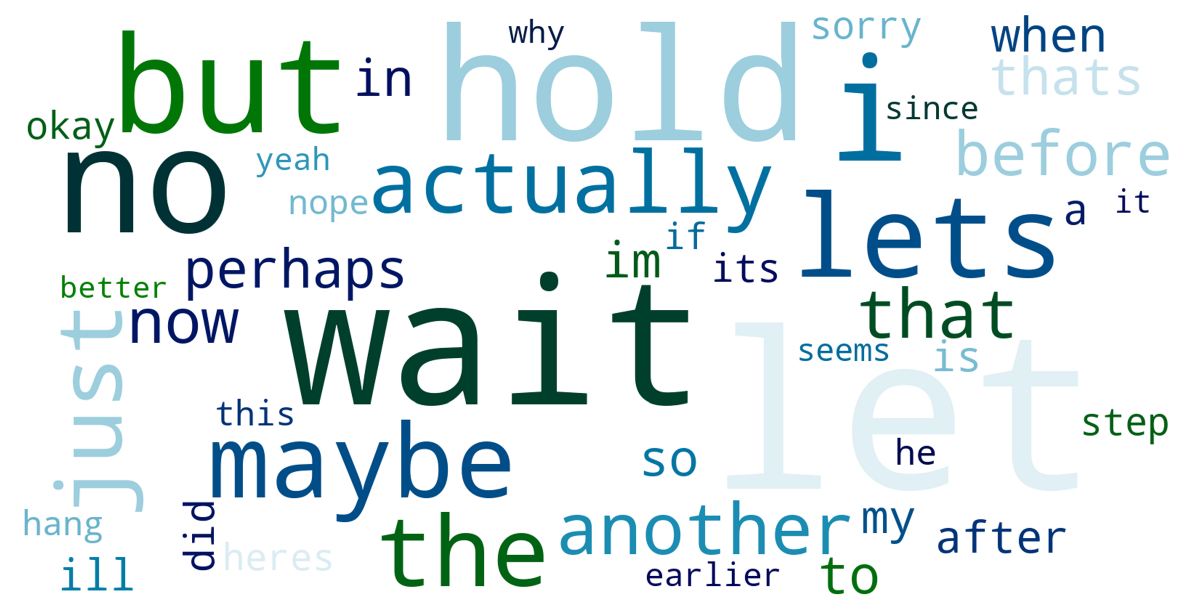}
 \caption{Word cloud of the first token generated immediately after injecting the learned \texttt{<|continue-thinking|>} token, across all datasets from \texttt{DS-R1-Qwen-1.5B}.}
\label{fig:word_cloud}
\end{figure}

To better understand how the learned token influences the model’s reasoning process, we visualize in \autoref{fig:word_cloud} a word cloud showing the first word generated immediately after the \texttt{<|continue-thinking|>} token. The most common continuations often prompt the model to self-verify or reconsider its previous steps, indicating that the token effectively encourages reflective reasoning and backtracking.

\begin{figure}[h]
\captionsetup{singlelinecheck=false}
    \begin{thinkbox}
    
    \small
    \textbf{Question:} Lloyd has an egg farm. His chickens produce 252 eggs per day and he sells them for \$2 per dozen. How much does Lloyd make on eggs per week?
    
    \vspace{0.1cm}
    \noindent\ttfamily
    {\textcolor{red!70!black}{{<think>}}} {\color{darkblue}First, I need to calculate the daily revenue from egg sales. [...] Finally, to determine the weekly revenue, I'll multiply the daily revenue by the number of days in a week, which is 7. So, \$1,134 multiplied by 7 equals \$7,938.
    }
    
    \vspace{0.1cm}
    \noindent
    {\textcolor{red!70!black}{{<|continue\_thinking|>}}} {\textcolor{green!40!black}{no, let me try that again. [...] Therefore, the correct answer is \$294 per week.
    }}

    \vspace{0.1cm}
    {\textcolor{red!70!black}{{</think>}}} Lloyd sells his eggs at \$2 per dozen. To determine his weekly revenue:
    \begin{enumerate}
        \item \textbf{Calculate weekly egg production}: - 252 eggs/day * 7 days = 1,764 eggs/week
    \end{enumerate} [...]
    
    \vspace{0.1cm}
    \noindent\textbf{Answer:} Lloyd makes \texttt{\textbackslash boxed\{294\}} dollars per week from his egg sales.
    \end{thinkbox}
  \caption{GSM8K reasoning trace from \texttt{DS-R1-Qwen-1.5B} demonstrating the positive impact of \texttt{<|continue\_thinking|>} token. Blue indicates the original reasoning, yielding an 
 incorrect answer of 7,938. Green shows the continuation after the special token was added, leading to the correct answer of 294. }
  \label{fig:example-reasoning-trace}
\end{figure}

The reasoning trace depicted in \autoref{fig:example-reasoning-trace}, taken from the GSM8K dataset, showcases how the \texttt{<|continue\_thinking|>} token influences the final answer. The blue portion highlights the model's initial reasoning, which leads to an incorrect result. However, the introduction of the \texttt{<|continue\_thinking|>} token prompts a re-examination of the solution, guiding the model to the correct conclusion. See~\autoref{sec:appendix-generated-examples} for the full reasoning traces and additional examples.

\section{Conclusions}
In this work, we have demonstrated that learning a dedicated \texttt{<|continue-thinking|>} token yields significant effectiveness, specifically in scenarios where the baseline budget forcing method already provides performance improvements. We introduced a training methodology for this token that exhibits promising generalization capabilities across different inference settings. Our analysis indicates that the observed performance gains are not primarily due to better adherence to output formatting, but rather stem from the elicited longer reasoning traces and a genuine enhancement in the model's underlying reasoning capabilities. Furthermore, this improvement remains relevant on average and conditionally across varying reasoning lengths, suggesting its utility even when generating shorter completions. While our method of learning a specialized \texttt{<|continue-thinking|>} token is relatively simple, practitioners can readily ascertain its potential benefit for their specific scenario by first employing the vanilla budget-forcing technique with a fixed token, such as ``Wait''; an observed performance increase with this baseline strongly suggests that training a dedicated \texttt{<|continue-thinking|>} token would be worthwhile. Finally, we emphasize the critical importance of rigorous evaluation for drawing meaningful conclusions and propose a refined evaluation scheme designed to mitigate some of the inherent limitations associated with relying solely on regex-based assessments.

\paragraph{Future Directions} One promising future direction involves exploring the efficacy of learning distinct \texttt{<|continue-thinking|>} tokens tailored to different positions within the generated sequence or investigating the benefits of learning these tokens jointly. For example, one can add a special token that will be used for the first forced continuation and a second new token that will be used for the second forced continuation.
More broadly, an exciting avenue is to train specialized tokens to enable additional forms of controllable reasoning, such as controlling answer length or enabling early termination.
Given our focus on the sequential approach to test-time scaling, another compelling future direction would be to explore the integration of our learned token methodology with parallel test-time scaling paradigms. Finally, extending the scope of this research to diverse domains, potentially even those lacking explicit verifiable rewards, such as in the context of LLM alignment, presents an intriguing area for future exploration. We believe the code provided with this paper will enable researchers to pursue these research directions efficiently.

\section{Limitations}
While our proposed method demonstrates promising results, it is subject to several limitations. First, as our analysis indicates, the effectiveness of the learned token appears to be contingent on the baseline performance of the budget forcing technique itself. If standard budget forcing does not yield improvements, our learned \texttt{<|continue-thinking|>} token is unlikely to provide a significant advantage. Second, the process of learning the token embedding necessitates a training phase, which is inherently more computationally demanding and requires access to the model's weights compared to simply employing fixed tokens. Finally, our method requires the addition of a new token to the model's vocabulary. This modification might not be feasible or permitted when utilizing LLMs through certain API interfaces, which often provide restricted access to the model's architecture, thus preventing vocabulary modifications. 

\section{Acknowledgments}
This research was supported by the European Union (ERC, SafetyBounds, 101163414). Views and opinions expressed are however those of the authors only and do not necessarily reflect those of the European Union or the European Research Council Executive Agency. Neither the European Union nor the granting authority can be held responsible for them. This research was also partially supported by the Israel Science Foundation (ISF grant 729/21). Y.~R. acknowledges additional support from the Career Advancement Fellowship at the Technion.

\bibliography{ref}

\clearpage
\appendix

\section{Training and Evaluation Parameters}
\autoref{tab:parameters} summarizes the full list of parameters we used for training and evaluation.

\begin{table}[ht]
\begin{center}
\begin{small}
\begin{adjustbox}{width=\linewidth}
\begin{tabular}{l p{5cm}}
\toprule
GRPO Parameters & \\ & \\
$G$ & 16 \\
$\varepsilon$ & $0.2$ \\
$\beta$ & $0$ \\
$\mu$ & $4$ \\ & \\
Training \\ Parameters & \\ & \\
Optimizer & AdamW \\
Learning rate &  $1e-4$ with cosine scheduler \\
Temperature & $0.9$ \\
Training and \\ Inference Prompt &  You are a helpful AI Assistant. First, think through the reasoning inside <think>...</think>. Then, always present the final answer in \textbackslash boxed\{\}   \\
\hline
Evaluation \\ Parameters & \\ & \\
Temperature  &  0.9 \\
LLM evaluator \\ prompt &  Given a math question, a correct answer, and a student's final answer (which may include explanations), determine if the correct answer appears in some form within the student's answer (ignoring trivial differences like formatting or wording). Output True if the student's answer is correct, otherwise output False. Output nothing else. \\ & \\
& START-OF-QUESTION \\ & \\
& question: \%(question)s \\ & \\
& END-OF-QUESTION \\ & \\
& correct answer: \%(correct\_answer)s \\ & \\
& student's solution: \\ & \\
& START-OF-STUDENT-SOLUTION \\ & \\
& \%(student\_solution)s \\ & \\
& END-OF-STUDENT-SOLUTION \\ & \\
& Output True if the student's solution equivalent the correct answer and False otherwise.
 \\
\bottomrule
\end{tabular}
\end{adjustbox}
\end{small}
\end{center}
\caption{Full list of training and evaluation parameters}
\label{tab:parameters}
\end{table}

\section{Artifacts Used}
The following datasets, software libraries and models were used during this research, all artifacts were used in accordance with their respective licenses.
\begin{itemize}
    \item \textbf{Datasets}: \texttt{DeepScaleR-Preview} Dataset, licensed under the MIT license.\footnote{https://hf.co/datasets/agentica-org/DeepScaleR-Preview-Dataset} \texttt{GSM8K-Platinum}, licensed under CC BY-SA 4.0.\footnote{https://hf.co/datasets/madrylab/gsm8k-platinum} \texttt{AIME24}, licensed under Apache-2.0,\footnote{ https://hf.co/datasets/simplescaling/aime24\_nofigures} and AIME25.\footnote{https://hf.co/datasets/math-ai/aime25}

    \item \textbf{Models}: \texttt{DeepSeek-R1-Distill-Qwen-1.5B}, \texttt{DeepSeek-R1-Distill-Qwen-7B} and \texttt{DeepSeek-R1-Distill-Llama-8B} licensed under the MIT license.\footnote{https://hf.co/deepseek-ai/DeepSeek-R1-Distill-Qwen-1.5B}\footnote{https://hf.co/deepseek-ai/DeepSeek-R1-Distill-Qwen-7B}\footnote{https://hf.co/deepseek-ai/DeepSeek-R1-Distill-Llama-8B}  \texttt{Qwen2.5-7B-Instruct}, licensed under the Apache-2.0  license.\footnote{https://hf.co/Qwen/Qwen2.5-7B-Instruct}

    \item \textbf{Software Packages}: LM-evaluation-harness licensed under the MIT license.\footnote{https://github.com/EleutherAI/lm-evaluation-harness}  vLLM licensed under the Apache-2.0.\footnote{https://github.com/vllm-project/vllm} Open r1, licensed under the Apache 2.0 license,\footnote{https://github.com/huggingface/open-r1} and trl, licensed under Apache 2.0.\footnote{https://github.com/huggingface/trl}
\end{itemize}

\section{Full Experimental Results}
We provide the complete set of results for all our experiments. \autoref{tab:results-full-1.5b}, \autoref{tab:results-full-7b}, and \autoref{tab:results-full-llama-8b} show the complete set of results when using an LLM evaluator in all configurations ($C_{\text{max}} = 1,2,3$, $B_{\text{max}} = 8192, 9216$). \autoref{tab:results_no_llm_full-1.5b} and  \autoref{tab:results_no_llm_full-7b} show the complete set of results for regex-only evaluation. \autoref{tab:probabilities-full} reports, for each configuration, the probability that the learned token yields a correct answer when the baseline does not, and the probability that the baseline yields a correct answer when the learned token does not.

\begin{table*}[ht]
\begin{small}
\begin{tabular}{l l l l l l l} 
 \toprule
  Dataset  & Baseline  & Alternatively & Hmm & Wait & Critique & Learned \\ 
 & (w/o BF) & & & & &\\ 
  \midrule
 $C_{\text{max}} \!= \!1$, $B_{\text{max}} \!=\! 8192$ & & & & & &\\
 AIME24  & 23.44 $\pm$ 0.73 & 23.07 $\pm$ 0.72 & 23.07 $\pm$ 0.70 & 23.18 $\pm$ 0.69 & 23.33  $\pm$ 0.51 & 23.07 $\pm$ 0.71 \\ 
 AIME25  & 21.82 $\pm$ 0.67 & 21.98 $\pm$ 0.56 & 22.08 $\pm$ 0.53 & 22.6 $\pm$ 0.51 & 22.76  $\pm$ 0.65 & 21.77 $\pm$ 0.55 \\
 GSM8K & 78.41 $\pm$ 0.25  & 78.43 $\pm$ 0.30 & 79.09 $\pm$ 0.42 & 79.09 $\pm$ 0.54 & 76.76  $\pm$ 0.24  & \textbf{81.39}$ \pm$ 0.40 \\
 MATH500 & 79.43 $\pm$ 0.26 & 80.26 $\pm$ 0.36 & 80.40 $\pm$ 0.29 & 81.09 $\pm$ 0.32  & 80.35  $\pm$ 0.36 & \textbf{82.00} $\pm$ 0.29  \\
 \hline
 $C_{\text{max}} \!= \!2$, $B_{\text{max}}\! = \!8192$ & & & & & &\\
 AIME24 & 23.44 $\pm$ 0.73  & 23.54 $\pm$ 0.82 & 23.85 $\pm$ 0.76 & 24.06 $\pm$ 0.76  & 23.28  $\pm$ 0.62 & 24.53 $\pm$ 0.75 \\ 
 AIME25 & 21.88 $\pm$ 0.67  & 21.72 $\pm$ 0.66 & 21.61 $\pm$ 0.67 & 21.98 $\pm$ 0.68  &  22.24  $\pm$ 0.63 & 22.34 $\pm$ 0.71 \\
 GSM8K  & 78.41 $\pm$ 0.25 & 78.58 $\pm$ 0.34 & 79.09 $\pm$ 0.28 & 79.71 $\pm$ 0.31  & 76.80  $\pm$ 0.46  & \textbf{82.63} $\pm$ 0.20\\
 MATH500 & 79.43 $\pm$ 0.26 & 80.00 $\pm$ 0.19 & 80.28 $\pm$ 0.29 & 80.36 $\pm$ 0.27  & 79.62  $\pm$ 0.31 & \textbf{81.67} $\pm$ 0.21 \\ 
 \hline
 $C_{\text{max}} \!= \!3$, $B_{\text{max}} \!=\! 8192$ & & & & & \\
AIME24 & 23.44 $\pm$ 0.73  & 23.02 $\pm$ 0.58  & 23.12 $\pm$ 0.62  & 23.49 $\pm$ 0.54  & 23.49  $\pm$ 0.75  & 23.33 $\pm$ 0.73  \\
AIME25 & 21.82 $\pm$ 0.67  & 22.08 $\pm$ 0.65  & 21.67 $\pm$ 0.68  & 22.03 $\pm$ 0.61  & 22.86  $\pm$ 0.60 & 22.03 $\pm$ 0.60  \\
GSM8K & 78.41 $\pm$ 0.25  & 79.00 $\pm$ 0.13  & 79.65 $\pm$ 0.26  & 80.16 $\pm$ 0.32   & 76.58  $\pm$ 0.16  & \textbf{83.17}  $\pm$ 0.32  \\
MATH500 & 79.42 $\pm$ 0.26  & 79.60 $\pm$ 0.32  & 80.71 $\pm$ 0.30  & 80.86 $\pm$ 0.18  & 80.24  $\pm$ 0.29   & \textbf{82.29}  $\pm$ 0.24  \\
 \hline
 $C_{\text{max}} \!= \!1$, $B_{\text{max}} \!=\! 9216$ & & & & & \\
 AIME24 & 23.85 $\pm$ 0.66  & 23.28 $\pm$ 0.63 & 24.17 $\pm$ 0.63 & 23.91 $\pm$ 0.62 & 23.59  $\pm$ 0.66 & 24.01 $\pm$ 0.61 \\ 
 AIME25 & 23.33 $\pm$ 0.52  & 23.33 $\pm$ 0.55 & 23.80 $\pm$ 0.60 & 23.70 $\pm$ 0.56 & 22.24  $\pm$ 0.55 & 23.28 $\pm$ 0.53 \\
 GSM8K  & 78.08 $\pm$ 0.36 & 78.83 $\pm$ 0.35 & 78.91 $\pm$ 0.33 & 79.11 $\pm$ 0.42 & 76.28  $\pm$ 0.43 & \textbf{81.67} $\pm$ 0.47\\
 MATH500  & 79.88 $\pm$ 0.33 & 80.09 $\pm$ 0.33 & 80.69 $\pm$ 0.32 & 80.82 $\pm$ 0.30 & 80.29  $\pm$ 0.24& 81.46 $\pm$ 0.39\\
  BBH & 45.71 $\pm$ 0.60  & 46.24 $\pm$ 0.60  & 46.53 $\pm$ 0.64  & 47.79 $\pm$ 0.70 & 45.91  $\pm$ 0.77  & 47.07  $\pm$ 0.66  \\
 \hline
 $C_{\text{max}} \!=\! 2$, $B_{\text{max}} \!=\! 9216$ & & & & & \\
 AIME24 & 23.85 $\pm$ 0.66  & 23.59 $\pm$ 0.68 & 24.37 $\pm$ 0.68 & 24.06 $\pm$ 0.75 & 22.86  $\pm$ 0.75  & 24.06 $\pm$ 0.69 \\ 
 AIME25 & 23.33 $\pm$ 0.52  & 23.39 $\pm$ 0.51 & 23.75 $\pm$ 0.53 & 23.54 $\pm$ 0.51 & 22.55 $\pm$ 0.52  & 23.07 $\pm$ 0.49 \\
 GSM8K  & 78.08 $\pm$ 0.36 & 78.96 $\pm$ 0.21 & 79.67 $\pm$ 0.35 & 79.86 $\pm$ 0.39 & 76.92  $\pm$ 0.45   & \textbf{82.53} $\pm$ 0.48\\
 MATH500 & 79.88 $\pm$ 0.33 & 80.14 $\pm$ 0.38 & 81.19 $\pm$ 0.31 & 81.21 $\pm$ 0.28 & 79.14  $\pm$ 0.29   & \textbf{82.30} $\pm$ 0.34 \\
 BBH & 45.71 $\pm$ 0.60  & 46.58 $\pm$ 0.50  & 46.78 $\pm$ 0.76  & 47.36 $\pm$ 0.62 & 45.97  $\pm$ 0.62  & 47.39  $\pm$ 0.57  \\
 \hline
 $C_{\text{max}} \!= \!3$, $B_{\text{max}}\! =\! 9216$ & & & & & \\
 AIME24 & 23.85 $\pm$ 0.66  & 23.49 $\pm$ 0.69  & 24.48 $\pm$ 0.63  & 23.91 $\pm$ 0.70  & 23.12  $\pm$ 0.71  & 24.06 $\pm$ 0.68  \\
AIME25 & 23.33 $\pm$ 0.52  & 23.80 $\pm$ 0.56  & 23.85 $\pm$ 0.53  & 23.49 $\pm$ 0.49  & 22.45  $\pm$ 0.49  & 22.40 $\pm$ 0.44  \\
GSM8K & 78.08 $\pm$ 0.36  & 78.83 $\pm$ 0.34  & 79.47 $\pm$ 0.34  & 80.14 $\pm$ 0.24  & 76.10 $\pm$ 0.59 & \textbf{83.68}  $\pm$ 0.38  \\
MATH500 & 79.88 $\pm$ 0.33  & 80.33 $\pm$ 0.35  & 80.84 $\pm$ 0.39  & 81.22 $\pm$ 0.30 & 79.49  $\pm$ 0.32  & \textbf{82.51}  $\pm$ 0.34  \\
BBH & 45.71 $\pm$ 0.60  & 46.75 $\pm$ 0.43  & 47.06 $\pm$ 0.60  & 47.58 $\pm$ 0.52 & 46.00  $\pm$ 0.72  & 47.70  $\pm$ 0.62  \\
 \bottomrule
\end{tabular}
\end{small}
\caption{Accuracy (pass@1) results for \texttt{DeepSeek-R1-Distill-Qwen-1.5B} with different token budget limits $B_{\text{max}}$ and different numbers of forced thinking continuations $C_{\text{max}}$. Results are obtained via regex-based evaluation and an LLM evaluator if the model fails to generate an answer in the correct format.}
\label{tab:results-full-1.5b}
\end{table*}

\begin{table*}[ht]
\begin{small}
\begin{tabular}{l l l l l l l} 
 \toprule
  Dataset  & Baseline  & Alternatively & Hmm & Wait & Critique & Learned \\ 
 & (w/o BF) & & & & &\\ 
  \midrule
 $C_{\text{max}}  \!= \! 1$, $B_{\text{max}}  \!= \! 8192$ \\
 AIME24 & 44.90 $\pm$ 0.75  & 45.16 $\pm$ 0.79  & 45.10 $\pm$ 0.75  & 45.57 $\pm$ 0.77  & 46.00 $\pm$ 0.82  & 46.20 $\pm$ 0.82  \\
 AIME25 & 35.83 $\pm$ 0.57  & 35.78 $\pm$ 0.60  & 35.36 $\pm$ 0.58  & 35.78 $\pm$ 0.66  & 35.83 $\pm$ 0.57  & 35.89 $\pm$ 0.59  \\
 GSM8K & 88.60 $\pm$ 0.35  & 88.50 $\pm$ 0.31  & 90.56 $\pm$ 0.20  & 89.74 $\pm$ 0.12  & 89.23 $\pm$ 0.28  & \textbf{94.60}  $\pm$ 0.22  \\
 MATH500 & 89.39 $\pm$ 0.26  & 89.09 $\pm$ 0.29  & 90.55 $\pm$ 0.25  & 90.34 $\pm$ 0.18  & 89.20 $\pm$ 0.27  & \textbf{92.64}  $\pm$ 0.24  \\
 \hline
 $C_{\text{max}}  \!= \! 2$, $B_{\text{max}}  \!= \! 8192$ \\
 AIME24 & 44.90 $\pm$ 0.75  & 45.00 $\pm$ 0.76  & 45.68 $\pm$ 0.79  & 45.78 $\pm$ 0.82  & 46.20 $\pm$ 0.84  & 45.63 $\pm$ 0.84  \\
 AIME25 & 35.83 $\pm$ 0.57  & 36.04 $\pm$ 0.64  & 35.99 $\pm$ 0.67  & 35.68 $\pm$ 0.67  & 35.00 $\pm$ 0.54  & 35.94 $\pm$ 0.60  \\
 GSM8K & 88.60 $\pm$ 0.35  & 89.16 $\pm$ 0.17  & 91.33 $\pm$ 0.21  & 90.89 $\pm$ 0.15  & 88.68 $\pm$ 0.23  & \textbf{95.00}  $\pm$ 0.16  \\
MATH500 & 89.39 $\pm$ 0.26  & 89.18 $\pm$ 0.26  & 91.16 $\pm$ 0.20  & 91.19 $\pm$ 0.21  & 89.14 $\pm$ 0.28  & \textbf{92.67}  $\pm$ 0.24  \\
 \hline
 $C_{\text{max}}  \!=  \!3$, $B_{\text{max}}  \!= \! 8192$ \\
AIME24 & 44.90 $\pm$ 0.75  & 45.26 $\pm$ 0.82  & 45.36 $\pm$ 0.86  & 46.46 $\pm$ 0.79  & 45.78 $\pm$ 0.78  & 45.63 $\pm$ 0.80  \\

AIME25 & 35.83 $\pm$ 0.57  & 35.83 $\pm$ 0.54  & 36.04 $\pm$ 0.63  & 36.35 $\pm$ 0.57  & 35.73 $\pm$ 0.63  & 35.99 $\pm$ 0.56  \\

GSM8K & 88.60 $\pm$ 0.35  & 89.15 $\pm$ 0.22  & 91.85 $\pm$ 0.24  & 91.43 $\pm$ 0.32  & 88.72 $\pm$ 0.25  & \textbf{95.18}  $\pm$ 0.17  \\

MATH500 & 89.39 $\pm$ 0.26  & 89.26 $\pm$ 0.26  & 91.78 $\pm$ 0.28  & 91.75 $\pm$ 0.18  & 88.85 $\pm$ 0.26  & \textbf{92.64}  $\pm$ 0.22  \\
 \hline
 $C_{\text{max}}  \!= \! 1$, $B_{\text{max}}  \!= \! 9216$ \\
AIME24 & 46.09 $\pm$ 0.80  & 46.77 $\pm$ 0.77  & 46.61 $\pm$ 0.77  & 47.08 $\pm$ 0.77  & 46.61 $\pm$ 0.77  & 46.25 $\pm$ 0.80  \\

AIME25 & 36.41 $\pm$ 0.66  & 35.78 $\pm$ 0.58  & 36.15 $\pm$ 0.55  & 36.35 $\pm$ 0.64  & 34.74 $\pm$ 0.62  & 36.46 $\pm$ 0.63  \\

GSM8K & 88.81 $\pm$ 0.30  & 88.59 $\pm$ 0.20  & 90.58 $\pm$ 0.25  & 90.28 $\pm$ 0.25  & 89.07 $\pm$ 0.10  & \textbf{94.40}  $\pm$ 0.29  \\

MATH500 & 89.41 $\pm$ 0.30  & 89.28 $\pm$ 0.19  & 91.00 $\pm$ 0.20  & 90.61 $\pm$ 0.25  & 89.44 $\pm$ 0.23  & \textbf{92.90}  $\pm$ 0.20  \\
BBH & 71.05 $\pm$ 0.75  & 71.70 $\pm$ 0.74  & 72.36 $\pm$ 0.65  & 72.55 $\pm$ 0.64 & 71.22  $\pm$ 0.72  & 73.91  $\pm$ 0.78  \\
 \hline
 $C_{\text{max}}  \!= \! 2$, $B_{\text{max}}  \!= \! 9216$ \\
AIME24 & 46.09 $\pm$ 0.80  & 46.41 $\pm$ 0.73  & 46.72 $\pm$ 0.75  & 46.88 $\pm$ 0.73  & 46.56 $\pm$ 0.75  & 46.46 $\pm$ 0.81  \\

AIME25 & 36.41 $\pm$ 0.66  & 36.61 $\pm$ 0.60  & 36.46 $\pm$ 0.64  & 36.30 $\pm$ 0.65  & 37.34 $\pm$ 0.66  & 35.73 $\pm$ 0.62  \\

GSM8K & 88.81 $\pm$ 0.30  & 88.92 $\pm$ 0.39  & 91.08 $\pm$ 0.26  & 91.05 $\pm$ 0.38  & 89.18 $\pm$ 0.16  & \textbf{94.69}  $\pm$ 0.21  \\

MATH500 & 89.41 $\pm$ 0.30  & 89.47 $\pm$ 0.24  & 91.53 $\pm$ 0.19  & 91.70 $\pm$ 0.22  & 89.16 $\pm$ 0.25  & \textbf{93.04}  $\pm$ 0.17  \\
BBH & 71.05 $\pm$ 0.75  & 72.33 $\pm$ 0.43  & 72.58 $\pm$ 0.68  & 73.36 $\pm$ 0.50 & 71.21  $\pm$ 0.67  & \textbf{74.53}  $\pm$ 0.62  \\
 \hline
 $C_{\text{max}}  \!= \! 3$, $B_{\text{max}}  \!= \! 9216$ \\
AIME24 & 46.09 $\pm$ 0.80  & 46.46 $\pm$ 0.73  & 47.24 $\pm$ 0.71  & 47.19 $\pm$ 0.77  & 46.41 $\pm$ 0.75  & 46.82 $\pm$ 0.76  \\
AIME25 & 36.41 $\pm$ 0.66  & 36.51 $\pm$ 0.57  & 36.51 $\pm$ 0.61  & 36.77 $\pm$ 0.63  & 37.29 $\pm$ 0.71  & 36.09 $\pm$ 0.58  \\
GSM8K & 88.81 $\pm$ 0.30  & 89.27 $\pm$ 0.27  & 91.23 $\pm$ 0.26  & 91.74 $\pm$ 0.34  & 88.89 $\pm$ 0.39  & \textbf{94.91}  $\pm$ 0.18  \\
MATH500 & 89.39 $\pm$ 0.31  & 89.84 $\pm$ 0.27  & 91.74 $\pm$ 0.19  & 92.17 $\pm$ 0.17  & 89.06 $\pm$ 0.27  & \textbf{92.89}  $\pm$ 0.20  \\
BBH & 71.05 $\pm$ 0.75  & 72.53 $\pm$ 0.72  & 72.91 $\pm$ 0.63  & 73.85 $\pm$ 0.51 & 71.01  $\pm$ 0.76  & 74.75  $\pm$ 0.60  \\
 \bottomrule
\end{tabular}
\end{small}
\caption{Accuracy (pass@1) results for \texttt{DeepSeek-R1-Distill-Qwen-7B} with different token budget limits $B_{\text{max}}$ and different numbers of forced thinking continuations $C_{\text{max}}$. Results are obtained via regex-based evaluation and an LLM evaluator if the model fails to generate an answer in the correct format.}
\label{tab:results-full-7b}
\end{table*}

\begin{table*}[ht]
\begin{small}
\begin{tabular}{l l l l l l l} 
 \toprule
  Dataset  & Baseline  & Alternatively & Hmm & Wait & Critique & Learned \\ 
 & (w/o BF) & & & & &\\ 
  \midrule
 $C_{\text{max}}  \!= \! 1$ \\
AIME24	& 35.68 $\pm$ 0.78	& 36.15 $\pm$ 0.78	& 38.12 $\pm$ 0.75	& 37.34 $\pm$ 0.75	& 36.56 $\pm$ 0.70	& \textbf{41.61} $\pm$ 0.82 \\
AIME25	& 29.32 $\pm$ 0.61	& 29.74 $\pm$ 0.58	& 29.27 $\pm$ 0.60	& 29.84 $\pm$ 0.61	& 29.69 $\pm$ 0.64	& 30.05 $\pm$ 0.66 \\
GSM8K	& 78.43 $\pm$ 0.40	& 78.22 $\pm$ 0.41	& 83.15 $\pm$ 0.40	& 81.28 $\pm$ 0.34	& 78.04 $\pm$ 0.56	& \textbf{93.47} $\pm$ 0.12 \\
MATH500	& 79.56 $\pm$ 0.31	& 80.64 $\pm$ 0.43	& 84.34 $\pm$ 0.28	& 83.41 $\pm$ 0.34	& 80.19 $\pm$ 0.35	& \textbf{89.71} $\pm$ 0.25 \\
BBH & 79.73 $\pm$ 0.37  & 80.58 $\pm$ 0.36  & 80.92 $\pm$ 0.44  & 80.98 $\pm$ 0.37 & 79.92  $\pm$ 0.40  & \textbf{83.25}  $\pm$ 0.24  \\
 \midrule
 $C_{\text{max}}  \!= \! 2$ \\
AIME24	& 35.68 $\pm$ 0.78	& 36.67 $\pm$ 0.78	& 37.71 $\pm$ 0.77	& 37.76 $\pm$ 0.86	& 36.15 $\pm$ 0.74	& \textbf{41.87} $\pm$ 0.87 \\
AIME25	& 29.32 $\pm$ 0.61	& 29.43 $\pm$ 0.56	& 29.06 $\pm$ 0.57	& 30.31 $\pm$ 0.60	& 29.48 $\pm$ 0.57	& 30.57 $\pm$ 0.61 \\ 
GSM8K	& 78.43 $\pm$ 0.40	& 79.56 $\pm$ 0.41	& 84.12 $\pm$ 0.28	& 83.82 $\pm$ 0.28	& 77.93 $\pm$ 0.46	& \textbf{93.73} $\pm$ 0.07 \\
MATH500	& 79.56 $\pm$ 0.31	& 81.09 $\pm$ 0.36	& 85.40 $\pm$ 0.24	& 85.11 $\pm$ 0.37	& 80.34 $\pm$ 0.26	& \textbf{89.91} $\pm$ 0.24 \\
BBH & 79.73 $\pm$ 0.37  & 81.01 $\pm$ 0.28  & 81.43 $\pm$ 0.47  & 81.71 $\pm$ 0.34 & 80.31  $\pm$ 0.29  & \textbf{83.61}  $\pm$ 0.32  \\
 \midrule
 $C_{\text{max}}  \!= \! 3$ \\
AIME24	& 35.68 $\pm$ 0.78	& 36.93 $\pm$ 0.75	& 38.23 $\pm$ 0.73	& 38.39 $\pm$ 0.79	& 35.73 $\pm$ 0.71	& \textbf{41.30} $\pm$ 0.91 \\
AIME25	& 29.32 $\pm$ 0.61	& 29.64 $\pm$ 0.57	& 30.10 $\pm$ 0.61	& 30.26 $\pm$ 0.64	& 29.90 $\pm$ 0.58	& 30.26 $\pm$ 0.65 \\
GSM8K	& 78.43 $\pm$ 0.40	& 79.82 $\pm$ 0.41	& 85.22 $\pm$ 0.22	& 84.90 $\pm$ 0.43	& 77.79 $\pm$ 0.65	& \textbf{93.78} $\pm$ 0.10 \\
MATH500	& 79.56 $\pm$ 0.31	& 81.86 $\pm$ 0.39	& 86.21 $\pm$ 0.32	& 86.01 $\pm$ 0.32	& 80.27 $\pm$ 0.40	& \textbf{90.04} $\pm$ 0.25 \\
BBH & 79.73 $\pm$ 0.37  & 81.32 $\pm$ 0.30  & 81.79 $\pm$ 0.46  & 81.97 $\pm$ 0.32 & 80.26  $\pm$ 0.42  & \textbf{83.69}  $\pm$ 0.37  \\
 \bottomrule
\end{tabular}
\end{small}
\caption{Accuracy (pass@1) results for \texttt{DeepSeek-R1-Distill-Llama-8B} with $B_{\text{max}}  \!= \! 9216$. Results are obtained via regex-based evaluation and an LLM evaluator if the model fails to generate an answer in the correct format.}
\label{tab:results-full-llama-8b}
\end{table*}

\begin{table*}[ht]
\begin{center}
\begin{small}
\begin{adjustbox}{width=\linewidth}
\begin{tabular}{l l l l l l l} 
 \toprule
 Dataset  & Baseline  & Alternatively & Hmm & Wait & Critique & Learned \\ 
 & (w/o BF)\\
  \midrule
 $C_{\text{max}}= 1$\\ $B_{\text{max}} = 8192$ & & & & & \\
 AIME24 & 22.86 $\pm$ 0.75 (81) & 22.71 $\pm$ 0.74 (81) & 22.34 $\pm$ 0.68 (81) & 22.60 $\pm$ 0.71 (82) & 22.50  $\pm$ 0.53 (80) & 22.66 $\pm$ 0.69 (82)  \\ 
 AIME25 & 21.56 $\pm$ 0.63 (85) & 21.67 $\pm$ 0.55 (84) & 21.72 $\pm$ 0.51 (84) & 22.19 $\pm$ 0.52 (82) & 21.98  $\pm$ 0.66 (83) & 21.61 $\pm$ 0.55 (84) \\
 GSM8K & 66.43 $\pm$ 0.50 (84) & 64.57 $\pm$ 0.33  (81) & 65.01 $\pm$ 0.56 (81) & 64.27 $\pm$ 0.37 (80) &68.57  $\pm$ 0.40 (86) & \textbf{75.71} $\pm$ 0.31 (92)  \\
 MATH500 & 77.19 $\pm$ 0.30 (94) & 77.89 $\pm$ 0.34 (94) & 77.94 $\pm$ 0.23 (94) & 78.59 $\pm$ 0.29  (94)& 78.35  $\pm$ 0.34 (94) & \textbf{80.84} $\pm$ 0.27 (96)\\
 \hline
 $C_{\text{max}} = 2$\\ $B_{\text{max}} = 8192$ & & & & & \\
 AIME24 & 22.86 $\pm$ 0.75 (81) & 23.02 $\pm$ 0.83 (80) & 23.23 $\pm$ 0.78 (80) & 23.28 $\pm$ 0.75 (82) & 22.71  $\pm$ 0.61 (79) & 24.22 $\pm$ 0.76  (81) \\ 
 AIME25 & 21.56 $\pm$ 0.63 (85) & 21.41 $\pm$ 0.65 (84) & 21.15 $\pm$ 0.67 (84) & 21.25 $\pm$ 0.64 (84)& 21.82 $\pm$ 0.64 (84) & 21.82 $\pm$ 0.68 (83) \\
 GSM8K & 66.43 $\pm$ 0.50 (84) & 64.52 $\pm$ 0.16 (81) & 64.56 $\pm$ 0.16 (80) & 63.55 $\pm$ 0.36 (79) & 70.03  $\pm$ 0.59 (87)  & \textbf{78.37} $\pm$ 0.40 (94) \\
 MATH500 & 77.19 $\pm$ 0.30 (94) & 77.64 $\pm$ 0.22 (94)& 77.99 $\pm$ 0.30 (94)& 78.05 $\pm$ 0.26 (94)& 77.85  $\pm$ 0.31 (95) & \textbf{80.74} $\pm$ 0.18 (96)\\ 
 \hline
 $C_{\text{max}} = 3$\\ $B_{\text{max}} = 8192$ & & & & & \\
AIME24 & 22.86 $\pm$ 0.75 (80) & 22.55 $\pm$ 0.51 (79) & 22.45 $\pm$ 0.57 (79) & 22.92 $\pm$ 0.52 (81) & 22.92  $\pm$ 0.74 (81) & 22.92 $\pm$ 0.74 (81) \\
AIME25 & 21.56 $\pm$ 0.63 (85) & 21.82 $\pm$ 0.67 (83) & 21.41 $\pm$ 0.67 (85) & 21.72 $\pm$ 0.61 (83) & 22.50  $\pm$ 0.57 (83)  & 21.56 $\pm$ 0.58 (84) \\
GSM8K & 66.43 $\pm$ 0.50 (83) & 64.21 $\pm$ 0.46 (80) & 64.42 $\pm$ 0.29 (79) & 62.99 $\pm$ 0.43 (77) & 70.25  $\pm$ 0.07 (88) & \textbf{80.11}  $\pm$ 0.30 (95) \\
MATH500 & 77.19 $\pm$ 0.30 (94) & 77.45 $\pm$ 0.31 (94) & 78.54 $\pm$ 0.30 (94) & 78.26 $\pm$ 0.20 (93)  & 78.31  $\pm$ 0.23 (95) & \textbf{81.60}  $\pm$ 0.22 (96) \\
 \hline
 $C_{\text{max}} = 1$  \\ $B_{\text{max}}= 9216$ & & & & & \\
 AIME24 & 23.18 $\pm$ 0.64 (82) & 22.86 $\pm$ 0.62 (82) & 23.54 $\pm$ 0.65 (81) & 23.39 $\pm$ 0.60 (82) & 23.18  $\pm$ 0.64 (81) & 23.70 $\pm$ 0.62  (82) \\ 
 AIME25 & 23.18 $\pm$ 0.50 (86) & 23.13 $\pm$ 0.54 (87) & 23.59 $\pm$ 0.58 (87) & 23.33 $\pm$ 0.58 (86)& 22.03  $\pm$ 0.54 (86) & 23.18 $\pm$ 0.53 (86) \\
 GSM8K & 66.03 $\pm$ 0.40 (84) & 65.76 $\pm$ 0.42 (82) & 64.82 $\pm$ 0.52 (81) & 64.24 $\pm$ 0.60 (80) & 68.25  $\pm$ 0.51 (86) & \textbf{75.12} $\pm$ 0.33 (91) \\
 MATH500 & 77.56 $\pm$ 0.32 (95) & 77.78 $\pm$ 0.25 (95)& 78.29 $\pm$ 0.34 (94) & 78.15 $\pm$ 0.32 (94) & 78.46  $\pm$ 0.23 (95) & \textbf{80.50} $\pm$ 0.35 (96)\\
 \hline
 $C_{\text{max}} = 2$ \\ $B_{\text{max}} = 9216$ & & & & & \\
 AIME24 & 23.18 $\pm$ 0.64 (82) & 23.12 $\pm$ 0.67 (82) & 23.91 $\pm$ 0.67 (83) & 23.75 $\pm$ 0.74 (83) & 22.40  $\pm$ 0.71 (82) & 23.59 $\pm$ 0.67  (83) \\ 
 AIME25 & 23.18 $\pm$ 0.50 (86) & 23.02 $\pm$ 0.49 (85) & 23.23 $\pm$ 0.51 (86) & 23.23 $\pm$ 0.50 (87)& 22.14  $\pm$ 0.49 (85) & 22.80 $\pm$ 0.47 (86) \\
 GSM8K & 66.03 $\pm$ 0.40 (84) & 64.82 $\pm$ 0.25 (82) & 64.09 $\pm$ 0.38 (80) & 64.43 $\pm$ 0.48 (79) & 69.99  $\pm$ 0.44 (88) & \textbf{78.27} $\pm$ 0.58 (94) \\
 MATH500  & 77.56 $\pm$ 0.32 (95)& 77.97 $\pm$ 0.35 (95)& 78.88 $\pm$ 0.32 (95)& 78.83 $\pm$ 0.38 (95)& 77.56  $\pm$ 0.28 (95) & \textbf{81.37} $\pm$ 0.38 (97)\\
 \hline
 $C_{\text{max}} = 3$ \\ $B_{\text{max}} = 9216$ & & & & & \\
AIME24 & 23.18 $\pm$ 0.64 (82) & 23.39 $\pm$ 0.70 (81) & 24.06 $\pm$ 0.67 (82) & 23.28 $\pm$ 0.72 (82) & 22.60  $\pm$ 0.68 (82) & 23.39 $\pm$ 0.66 (83) \\
AIME25 & 23.18 $\pm$ 0.50 (85) & 23.44 $\pm$ 0.56 (87) & 23.49 $\pm$ 0.49 (86) & 23.12 $\pm$ 0.46 (86) & 22.14  $\pm$ 0.46 (86) & 22.40 $\pm$ 0.44 (86) \\
GSM8K & 66.03 $\pm$ 0.40 (83) & 63.79 $\pm$ 0.57 (80) & 64.27 $\pm$ 0.28 (79) & 62.94 $\pm$ 0.50 (77) & 69.35  $\pm$ 0.56 (87) & \textbf{80.87}  $\pm$ 0.42 (96) \\
MATH500 & 77.56 $\pm$ 0.32 (94) & 78.16 $\pm$ 0.39 (95) & 78.60 $\pm$ 0.40 (94) & 78.75 $\pm$ 0.32 (94) & 78.08  $\pm$ 0.28 (96) & \textbf{81.75}  $\pm$ 0.34 (96) \\
 \bottomrule
\end{tabular}
\end{adjustbox}
\end{small}
\end{center}
\caption{Accuracy (pass@1) results for \texttt{DeepSeek-R1-Distill-Qwen-1.5B} with different token budget limits $B_{\text{max}}$ and different numbers of forced thinking continuations $C_{\text{max}}$. Results are obtained via a regex-based evaluation only. The percentage of final answers enclosed in \texttt{\textbackslash boxed\{\}} is shown in parentheses.}
\label{tab:results_no_llm_full-1.5b}
\end{table*}

\begin{table*}[ht]
\begin{center}
\begin{small}
\begin{adjustbox}{width=\linewidth}
\begin{tabular}{l l l l l l l} 
 \toprule
 Dataset & Baseline (w/o BF) & Alternatively & Hmm & Wait & Critique & Learned \\  
  \midrule
 $C_{\text{max}}= 1$\\ $B_{\text{max}} = 8192$ & & & & & \\
AIME24 & 43.85 $\pm$ 0.77 (73) & 43.65 $\pm$ 0.77 (71) & 43.91 $\pm$ 0.77 (71) & 44.38 $\pm$ 0.80 (71) & 44.64 $\pm$ 0.82 (72) & 44.79 $\pm$ 0.79 (72) \\
AIME25 & 34.74 $\pm$ 0.58 (65) & 34.74 $\pm$ 0.60 (68) & 34.69 $\pm$ 0.57 (67) & 34.58 $\pm$ 0.57 (66) & 34.74 $\pm$ 0.62 (65) & 34.95 $\pm$ 0.56 (66) \\
GSM8K & 88.56 $\pm$ 0.34 (99) & 88.46 $\pm$ 0.30 (99) & 90.45 $\pm$ 0.19 (99) & 89.63 $\pm$ 0.13 (99) & 89.19 $\pm$ 0.27 (99) & \textbf{94.43}  $\pm$ 0.26 (99) \\
MATH500 & 88.72 $\pm$ 0.24 (97) & 88.55 $\pm$ 0.31 (97) & 90.06 $\pm$ 0.24 (97) & 89.91 $\pm$ 0.21 (97) & 88.54 $\pm$ 0.27 (97) & \textbf{92.20}  $\pm$ 0.24 (97) \\
 \hline
 $C_{\text{max}} = 2$\\ $B_{\text{max}} = 8192$ & & & & & \\
AIME24 & 43.85 $\pm$ 0.77 (73) & 44.32 $\pm$ 0.76 (73) & 44.53 $\pm$ 0.80 (73) & 44.53 $\pm$ 0.80 (72) & 44.84 $\pm$ 0.80 (71) & 44.48 $\pm$ 0.82 (71) \\
AIME25 & 34.74 $\pm$ 0.58 (65) & 34.90 $\pm$ 0.58 (67) & 34.90 $\pm$ 0.61 (65) & 34.74 $\pm$ 0.63 (66) & 34.17 $\pm$ 0.59 (65) & 34.69 $\pm$ 0.59 (66) \\
GSM8K & 88.56 $\pm$ 0.34 (99) & 89.07 $\pm$ 0.19 (99) & 91.03 $\pm$ 0.22 (99) & 90.65 $\pm$ 0.17 (99) & 88.61 $\pm$ 0.24 (99) & \textbf{94.91}  $\pm$ 0.16 (99) \\
MATH500 & 88.72 $\pm$ 0.24 (97) & 88.60 $\pm$ 0.27 (97) & 90.80 $\pm$ 0.20 (97) & 90.71 $\pm$ 0.21 (97) & 88.65 $\pm$ 0.29 (97) & \textbf{92.28}  $\pm$ 0.20 (97) \\
 \hline
 $C_{\text{max}} = 3$\\ $B_{\text{max}} = 8192$ & & & & & \\
AIME24 & 43.85 $\pm$ 0.77 (73) & 44.32 $\pm$ 0.84 (71) & 44.06 $\pm$ 0.81 (72) & 45.00 $\pm$ 0.79 (71) & 44.17 $\pm$ 0.78 (71) & 44.58 $\pm$ 0.78 (72) \\
AIME25 & 34.74 $\pm$ 0.58 (65) & 34.79 $\pm$ 0.57 (66) & 34.53 $\pm$ 0.64 (65) & 34.64 $\pm$ 0.59 (66) & 34.48 $\pm$ 0.66 (66) & 34.69 $\pm$ 0.57 (65) \\
GSM8K & 88.56 $\pm$ 0.34 (99) & 89.00 $\pm$ 0.24 (99) & 91.11 $\pm$ 0.29 (99) & 90.96 $\pm$ 0.28 (99) & 88.61 $\pm$ 0.24 (99) & \textbf{95.11}  $\pm$ 0.17 (99) \\
MATH500 & 88.72 $\pm$ 0.24 (97) & 88.64 $\pm$ 0.27 (97) & 91.13 $\pm$ 0.27 (97) & 91.36 $\pm$ 0.18 (97) & 88.29 $\pm$ 0.30 (97) & \textbf{92.24}  $\pm$ 0.20 (97) \\
 \hline
 $C_{\text{max}} = 1$  \\ $B_{\text{max}}= 9216$ & & & & & \\
AIME24 & 44.74 $\pm$ 0.77 (71) & 45.42 $\pm$ 0.79 (71) & 45.42 $\pm$ 0.78 (72) & 45.99 $\pm$ 0.72 (72) & 45.26 $\pm$ 0.77 (72) & 45.16 $\pm$ 0.79 (71) \\
AIME25 & 35.47 $\pm$ 0.62 (67) & 35.16 $\pm$ 0.61 (67) & 35.47 $\pm$ 0.55 (66) & 35.47 $\pm$ 0.61 (66) & 36.61 $\pm$ 0.68 (68) & 35.31 $\pm$ 0.56 (68) \\
GSM8K & 88.81 $\pm$ 0.30 (99) & 88.54 $\pm$ 0.19 (99) & 90.52 $\pm$ 0.24 (99) & 90.20 $\pm$ 0.24 (99) & 89.00 $\pm$ 0.10 (99) & \textbf{94.33}  $\pm$ 0.29 (99) \\
MATH500 & 88.94 $\pm$ 0.30 (97) & 88.75 $\pm$ 0.20 (97) & 90.68 $\pm$ 0.21 (97) & 90.18 $\pm$ 0.23 (97) & 88.86 $\pm$ 0.20 (97) & \textbf{92.58}  $\pm$ 0.20 (97) \\$\pm$ 0.20 (97) \\
 \hline
 $C_{\text{max}} = 2$ \\ $B_{\text{max}} = 9216$ & & & & & \\
AIME24 & 44.74 $\pm$ 0.77 (71) & 45.47 $\pm$ 0.78 (71) & 45.94 $\pm$ 0.78 (72) & 45.62 $\pm$ 0.79 (72) & 45.42 $\pm$ 0.77 (71) & 45.52 $\pm$ 0.80 (72) \\
AIME25 & 35.47 $\pm$ 0.62 (67) & 35.68 $\pm$ 0.59 (68) & 35.00 $\pm$ 0.62 (66) & 35.62 $\pm$ 0.66 (67) & 36.82 $\pm$ 0.67 (67) & 34.90 $\pm$ 0.60 (68) \\
GSM8K & 88.81 $\pm$ 0.30 (99) & 88.82 $\pm$ 0.39 (99) & 90.58 $\pm$ 0.29 (99) & 90.85 $\pm$ 0.39 (99) & 89.12 $\pm$ 0.14 (99) & \textbf{94.64}  $\pm$ 0.19 (99) \\
MATH500 & 88.94 $\pm$ 0.30 (97) & 88.90 $\pm$ 0.24 (97) & 91.08 $\pm$ 0.19 (97) & 91.24 $\pm$ 0.20 (97) & 88.74 $\pm$ 0.26 (97) & \textbf{92.65}  $\pm$ 0.18 (97) \\
 \hline
 $C_{\text{max}} = 3$ \\ $B_{\text{max}} = 9216$ & & & & & \\
AIME24 & 44.74 $\pm$ 0.77 (71) & 45.52 $\pm$ 0.72 (72) & 46.15 $\pm$ 0.72 (71) & 46.30 $\pm$ 0.74 (73) & 45.21 $\pm$ 0.80 (73) & 45.94 $\pm$ 0.79 (72) \\
AIME25 & 35.47 $\pm$ 0.62 (67) & 35.47 $\pm$ 0.57 (66) & 35.52 $\pm$ 0.62 (67) & 35.73 $\pm$ 0.64 (66) & 36.41 $\pm$ 0.68 (67) & 35.16 $\pm$ 0.58 (66) \\
GSM8K & 88.81 $\pm$ 0.30 (99) & 89.15 $\pm$ 0.25 (99) & 90.45 $\pm$ 0.24 (98) & 91.51 $\pm$ 0.35 (99) & 88.74 $\pm$ 0.38 (99) & \textbf{94.84}  $\pm$ 0.18 (99) \\
MATH500 & 88.94 $\pm$ 0.30 (97) & 89.42 $\pm$ 0.27 (97) & 91.20 $\pm$ 0.16 (97) & 91.78 $\pm$ 0.18 (97) & 88.70 $\pm$ 0.26 (97) & \textbf{92.62}  $\pm$ 0.20 (97) \\
 \bottomrule
\end{tabular}
\end{adjustbox}
\end{small}
\end{center}
\caption{Accuracy (pass@1) results for \texttt{DeepSeek-R1-Distill-Qwen-7B} with different token budget limits $B_{\text{max}}$ and different numbers of forced thinking continuations $C_{\text{max}}$. Results are obtained via a regex-based evaluation only. The percentage of final answers enclosed in \texttt{\textbackslash boxed\{\}} is shown in parentheses.}
\label{tab:results_no_llm_full-7b}
\end{table*}

\begin{table*}[ht]
\begin{center}
\begin{small}
\begin{tabular}{l l l l l} 
 \toprule
 Probabilities & AIME24 & AIME25 & GSM8K & MATH500 \\  
  \midrule
 $C_{\text{max}} = 1$, $B_{\text{max}} = 8192$ & & & & \\
 P(Learned \x, Baseline \tick) & 0.09 & 0.06 & 0.03 & 0.06 \\
 P(Learned \tick, Baseline \x) & 0.08 & 0.06 & 0.06 & 0.09 \\
 \hline
 $C_{\text{max}} = 2$, $B_{\text{max}} = 8192$ & & & &  \\
 P(Learned \x, Baseline \tick) & 0.01 & 0.01 & 0.03 & 0.02 \\
 P(Learned \tick, Baseline \x) & 0.02 & 0.02 & 0.07 & 0.04 \\
 \hline
 $C_{\text{max}} = 3$, $B_{\text{max}} = 8192$ & & & &  \\
 P(Learned \x, Baseline \tick) & 0.09 & 0.06 & 0.02 & 0.01 \\
 P(Learned \tick, Baseline \x) & 0.09 & 0.06 & 0.07 & 0.04 \\
 \hline
 $C_{\text{max}} = 1$, $B_{\text{max}} = 9216$ & & & &  \\
 P(Learned \x, Baseline \tick) & 0.02 & 0.01 & 0.03 & 0.02 \\
 P(Learned \tick, Baseline \x) & 0.02 & 0.01 & 0.06 & 0.03 \\
 \hline
 $C_{\text{max}} = 2$, $B_{\text{max}} = 9216$ & & & &  \\
 P(Learned \x, Baseline \tick) & 0.08 & 0.06 & 0.02 & 0.02 \\
 P(Learned \tick, Baseline \x) & 0.08 & 0.06 & 0.07 & 0.04 \\
 \hline
 $C_{\text{max}} = 3$, $B_{\text{max}} = 9216$ & & & &  \\
 P(Learned \x, Baseline \tick) & 0.09 & 0.06 & 0.02 & 0.02 \\
 P(Learned \tick, Baseline \x) & 0.09 & 0.06 & 0.07 & 0.04 \\
 \bottomrule
\end{tabular}
\end{small}
\end{center}
\caption{Probability of observing a correct answer from the learned token and an incorrect answer from the baseline and vice versa when using \texttt{DeepSeek-R1-Distill-Qwen-1.5B}.}
\label{tab:probabilities-full}
\end{table*}

\section{Generated Answers Examples}
\label{sec:appendix-generated-examples}
We include examples of reasoning traces of cases when using the learned token resulted in a correct answer while the baseline did not, and vice versa. In all figures, blue text indicates the original reasoning trace that is common for both the learned model and the baseline model and green text indicates the reasoning trace that was generated after a forced continuation. \autoref{fig:example-learned-correct-full} and \autoref{fig:example-baseline-incorrect-full} show full reasoning traces for the learned model and the baseline model respectively on a question in which the baseline model was incorrect and adding the learned token allowed the model to continue reasoning through the problem and reach the correct answer. \autoref{fig:example-both-correct-learned-full} and \autoref{fig:both-correct-baseline--full} show full reasoning traces for the learned model and the baseline model, respectively, on a question in which both models were correct. Note that in this case, the learned token only adds a negligible number of tokens to the answer.

\begin{figure*}[t]
\captionsetup{justification=raggedright,singlelinecheck=false}
    \begin{thinkboxtitle}{Learned token}
    \small
    \textbf{Question:} Lloyd has an egg farm. His chickens produce 252 eggs per day and he sells them for \$2 per dozen. How much does Lloyd make on eggs per week?

    \vspace{0.1cm}
    {\textcolor{red!70!black}{{<think>}}}
    {\color{darkblue} First, I need to calculate the daily revenue from egg sales. Each chicken produces 252 eggs daily, so for 27 chickens, the total daily egg production is 252 multiplied by 27, which equals 6,804 eggs. Next, I'll convert the number of eggs into dozens since the selling price is given per dozen. There are 12 eggs in a dozen, so the total dozens of eggs sold daily are 6,804 divided by 12, resulting in 567 dozen. Now, I'll find out how much 567 dozen eggs sell for per day by multiplying the number of dozens by the selling price per dozen. That's 567 multiplied by \$2, which equals \$1,134. Finally, to determine the weekly revenue, I'll multiply the daily revenue by the number of days in a week, which is 7. So, \$1,134 multiplied by 7 equals \$7,938.
    }
    
    \vspace{0.1cm}
    \noindent
    {\textcolor{red!70!black}{{<|continue\_thinking|>}}} {\textcolor{green!40!black}{no, let me try that again. If I have 27 chickens each producing 252 eggs, that's 27*252=6,804 eggs per day. He sells them for \$2 per dozen,which is \$2/12 per egg. So to find daily revenue, it's 6,804/12 *2 = \$1,134. Then multiply by 7 to get weekly: 1,134*7= \$7,938. Wait, did I just repeat the same steps? Maybe I need to approach it differently. Alternatively, first calculate the daily revenue. 252 eggs produce 252/12=21 dozen. Each dozen sells for \$2, so 21*2=\$42 per day. Then, over a week (7 days), it's 42*7=\$294. But I'm confused earlier got \$7,938. Which one is correct? Let me double-check. 27 chickens at 252 eggs each is indeed 27*252=6,804 eggs. 6,804 divided by 12 is 567 dozen. 567*2= \$1,134 per day. Then 1,134*7= \$7,938. Oh, wait. I think I'm making a mistake in the calculation order. Let me clarify: Egg production: 27 chickens * 252 eggs/day = 6,804 eggs/day. Dollars per egg: \$2 per dozen, which is \$2/12= \$0.1667 per egg. Total revenue per day: 6,804 eggs/day * \$0.1667/egg $\approx$ \$1,134. Revenue per week: \$1,134/day *7= \$7,938. But when I calculated earlier, I got a different result. What's the mistake? Wait, 252 eggs per day is 252 divided by 12 is 21 dozen. 21*2= \$42 daily. 42*7= \$294. Now, why the discrepancy? Let's check 27*252. 252*27: 27*200=5,400, 27*52=1,404, total is 5,400+1,404=6,804. Yes, that's correct. So 27 chickens produce 6,804 eggs a day. At \$2 per dozen, 6,804/12=567 dozen, 567*2= \$1,134. 1,134*7= \$7,938. But why is another method giving a different answer? Wait, maybe I'm interpreting the question wrong. If there are 27 chickens, each producing 252 eggs a day, so total eggs per day are 27*252=6,804. To find revenue, I need to know how much they sell each day. If they sell each dozen for \$2, then per day revenue is 6,804/12*\$2=567*\$2= \$1,134. Over 7 days, \$1,134*7= \$7,938. Alternatively, 27 chickens, 252 eggs/day: total eggs per week. 252*7=1,764 eggs/week. Convert to dozens: 1,764/12=147 dozen. 147*2= \$294 per week. Ah, I see now. Depending on how you group the eggs, you get different per-week revenues. The confusion comes from whether the revenue is per dozen or per week. The question says "how much does Lloyd make on eggs per week." Since the chickens produce 252 eggs a day, and he sells them for \$2 per dozen, the method is to calculate how much he makes per dozen eggs multiplied by how many dozens he sells per week. So, per week, he sells 252*7=1,764 eggs. Divided by 12, that's 147 dozen. 147*2= \$294 per week. Therefore, the correct answer is \$294 per week.}}
    
    \vspace{0.1cm}
    {\textcolor{red!70!black}{{</think>}}} Lloyd sells his eggs at \$2 per dozen. To determine his weekly revenue:
    \begin{enumerate}
        \item \textbf{Calculate weekly egg production}: 252 eggs/day * 7 days = 1,764 eggs/week
        \item \textbf{Convert weekly eggs to dozens}: 1,764 eggs / 12 = 147 dozen
        \item \textbf{Calculate weekly revenue}: 147 dozen * \$2/dozen = \$294
    \end{enumerate}
    
    \vspace{0.1cm}
    \noindent\textbf{Answer:} Lloyd makes \texttt{\textbackslash boxed\{294\}} dollars per week from his egg sales.
    \end{thinkboxtitle}

  \caption{GSM8K reasoning trace  from \texttt{DeepSeek-R1-Distill-Qwen-1.5B} demonstrating the positive impact of the \texttt{<|continue\_thinking|>} token. Blue indicates the original reasoning, yielding an incorrect answer of 7,938. Green shows the continuation after the special token was added, leading to the correct answer of 294.}
  \label{fig:example-learned-correct-full}
\end{figure*}

\begin{figure*}[t]
\captionsetup{justification=raggedright,singlelinecheck=false}

    \begin{thinkboxtitle}{Baseline}
    \small
    \textbf{Question:} Lloyd has an egg farm. His chickens produce 252 eggs per day and he sells them for \$2 per dozen. How much does Lloyd make on eggs per week?

    \vspace{0.1cm}
    {\textcolor{red!70!black}{{<think>}}}
    {\color{darkblue} First, I need to calculate the daily revenue from egg sales. Each chicken produces 252 eggs daily, so for 27 chickens, the total daily egg production is 252 multiplied by 27, which equals 6,804 eggs. Next, I'll convert the number of eggs into dozens since the selling price is given per dozen. There are 12 eggs in a dozen, so the total dozens of eggs sold daily are 6,804 divided by 12, resulting in 567 dozen. Now, I'll find out how much 567 dozen eggs sell for per day by multiplying the number of dozens by the selling price per dozen. That's 567 multiplied by \$2, which equals \$1,134. Finally, to determine the weekly revenue, I'll multiply the daily revenue by the number of days in a week, which is 7. So, \$1,134 multiplied by 7 equals \$7,938.
    }
    
    \vspace{0.1cm}
    {\textcolor{red!70!black}{{</think>}}}
    \vspace{0.1cm}
    
    \textbf{Solution}
    \vspace{0.1cm}

To determine how much Lloyd makes on eggs per week, we'll break down the problem step by step.
\begin{enumerate}
    \item \textbf{Calculate the total number of eggs produced per day:}
    \[
   \text{Eggs per day} = \text{Number of chickens} \times \text{Eggs per chicken per day}
   \]
   \[
   \text{Eggs per day} = 27 \times 252 = 6,\!804 \text{ eggs}
   \]
   \item \textbf{Determine the number of dozens of eggs produced per day:}
   \[
   \text{Dozens per day} = \frac{\text{Eggs per day}}{12}
   \]
   \[
   \text{Dozens per day} = \frac{6,\!804}{12} = 567 \text{ dozen}
   \]
   \item \textbf{Calculate the daily revenue from selling eggs:}
   \[
   \text{Revenue per day} = \text{Dozens per day} \times \text{Selling price per dozen}
   \]
   \[
   \text{Revenue per day} = 567 \times 2 = 1,\!134 \text{ dollars}
   \]
   \item \textbf{Find the weekly revenue by multiplying daily revenue by the number of days in a week:}
   \[
   \text{Revenue per week} = \text{Revenue per day} \times 7
   \]
   \[
   \text{Revenue per week} = 1,\!134 \times 7 = 7,\!938 \text{ dollars}
   \]
\end{enumerate}

    \vspace{0.1cm}
    \noindent\textbf{Final Answer:}
        \Verb|\boxed{7,938 dollars}|

    \end{thinkboxtitle}

  \caption{GSM8K reasoning trace  from \texttt{DeepSeek-R1-Distill-Qwen-1.5B} of the baseline model for the same question as in \autoref{fig:example-learned-correct-full}. The final answer provided by the baseline model is incorrect, as opposed to the correct answer given in \autoref{fig:example-learned-correct-full}. }
  \label{fig:example-baseline-incorrect-full}
\end{figure*}

\begin{figure*}[t]
\captionsetup{justification=raggedright,singlelinecheck=false}

    \begin{thinkboxtitle}{Learned token}
    \small
    \textbf{Question:} Twenty tourists discovered 700 shells in a strip mall parking lot. They had divided into three groups, Alphas, The finders, and Gogetters to find as many shells as possible. If team Alphas found 40\% of the shells, and team The finders found 60\% of the remaining shells, how many shells did team Gogetters find?

    \vspace{0.1cm}
    {\textcolor{red!70!black}{{<think>}}}
    {\color{darkblue} Okay, let's try to figure out how many shells team Gogetters found. So, there are 700 shells in total, and 20 tourists split them into three groups: Alphas, The finders, and Gogetters. I need to find out how many each group found. First, team Alphas found 40\% of the shells. That sounds like a straightforward calculation. So, 40\% of 700. Let me write that down: 0.4 * 700. Hmm, 0.4 is like 40\%, so if I multiply 700 by 0.4, what do I get? Let me do that math. 0.4 times 700... Yeah, that's 280. So, team Alphas found 280 shells. 
    Now, after team Alphas, there are 700 - 280 = 420 shells left. That number is the remaining shells after Alphas' team found theirs. Next, team The finders found 60\% of these remaining shells. So, I need to calculate 60\% of 420. Alright, 0.6 times 420. Let me compute that. 0.6 * 400 is 240, and 0.6 * 20 is 12. So, 240 + 12 equals 252. Therefore, team The finders found 252 shells. Now, to find out how many shells team Gogetters found, I subtract the shells found by both Alphas and The finders from the total. So, 700 - 280 - 252. Let me do that step by step. First, 700 - 280 is 420, as we had before. Then, 420 - 252. Hmm, 420 minus 200 is 220, and then minus 52 more is 168. So, team Gogetters found 168 shells.
    Let me just double-check my calculations to make sure I didn't make a mistake. Alphas: 40\% of 700 is indeed 280. Subtracting that leaves 420. The finders take 60\% of 420, which is 252. Subtracting 252 from 420 gives 168 for Gogetters. That all adds up correctly. I think that's all. The Gogetters found 168 shells.
    }
    
    \vspace{0.1cm}
    \noindent
    {\textcolor{red!70!black}{{<|continue\_thinking|>}}} {\textcolor{green!40!black}{, and he's right.}}
    
    \vspace{0.1cm}
    {\textcolor{red!70!black}{{</think>}}} To determine how many shells team Gogetters found, follow these steps:
    \begin{enumerate}
        \item \textbf{Calculate the number of shells found by team Alphas:} 
        \[
            40\% \text{ of } 700 = 0.4 \times 700 = 280 \text{ shells}
        \]
        \item \textbf{Subtract the shells found by Alphas from the total:}:
         \[
            700 - 280 = 420 \text{ shells remaining}
        \]
        \item \textbf{Calculate the number of shells found by team The finders:}
         \[
            60\% \text{ of } 420 = 0.6 \times 420 = 252 \text{ shells}
        \]
        \item \textbf{Subtract the shells found by The finders from the remaining shells:}
         \[
            420 - 252 = 168 \text{ shells}
        \]
    \end{enumerate}
    
    \vspace{0.1cm}
    \noindent\textbf{Conclusion:} Team Gogetters found \Verb|\boxed{168}| shells.
    \end{thinkboxtitle}

  \caption{GSM8K reasoning trace  from \texttt{DeepSeek-R1-Distill-Qwen-1.5B} demonstrating that the \texttt{<|continue\_thinking|>} token does not generate many tokens when the model is confident. Blue indicates the original reasoning. Green shows the short continuation after the special token was added.}
  \label{fig:example-both-correct-learned-full}
\end{figure*}

\begin{figure*}[t]
\captionsetup{justification=raggedright,singlelinecheck=false}

    \begin{thinkboxtitle}{Baseline}
    \small
    \textbf{Question:} Twenty tourists discovered 700 shells in a strip mall parking lot. They had divided into three groups, Alphas, The finders, and Gogetters to find as many shells as possible. If team Alphas found 40\% of the shells, and team The finders found 60\% of the remaining shells, how many shells did team Gogetters find?

    \vspace{0.1cm}
    {\textcolor{red!70!black}{{<think>}}}
    {\color{darkblue} Okay, let's try to figure out how many shells team Gogetters found. So, there are 700 shells in total, and 20 tourists split them into three groups: Alphas, The finders, and Gogetters. I need to find out how many each group found. First, team Alphas found 40\% of the shells. That sounds like a straightforward calculation. So, 40\% of 700. Let me write that down: 0.4 * 700. Hmm, 0.4 is like 40\%, so if I multiply 700 by 0.4, what do I get? Let me do that math. 0.4 times 700... Yeah, that's 280. So, team Alphas found 280 shells. 
    Now, after team Alphas, there are 700 - 280 = 420 shells left. That number is the remaining shells after Alphas' team found theirs. Next, team The finders found 60\% of these remaining shells. So, I need to calculate 60\% of 420. Alright, 0.6 times 420. Let me compute that. 0.6 * 400 is 240, and 0.6 * 20 is 12. So, 240 + 12 equals 252. Therefore, team The finders found 252 shells. Now, to find out how many shells team Gogetters found, I subtract the shells found by both Alphas and The finders from the total. So, 700 - 280 - 252. Let me do that step by step. First, 700 - 280 is 420, as we had before. Then, 420 - 252. Hmm, 420 minus 200 is 220, and then minus 52 more is 168. So, team Gogetters found 168 shells.
    Let me just double-check my calculations to make sure I didn't make a mistake. Alphas: 40\% of 700 is indeed 280. Subtracting that leaves 420. The finders take 60\% of 420, which is 252. Subtracting 252 from 420 gives 168 for Gogetters. That all adds up correctly. I think that's all. The Gogetters found 168 shells.
    }
    
    \vspace{0.1cm}
    {\textcolor{red!70!black}{{</think>}}}
    \vspace{0.1cm}

There are 700 shells in total. Team Alphas found 40

\[
0.4 \times 700 = 280 \text{ shells}
\]

The remaining shells after Alphas' team are:

\[
700 - 280 = 420 \text{ shells}
\]

Team The finders found 60

\[
0.6 \times 420 = 252 \text{ shells}
\]

Finally, the number of shells found by team Gogetters is:

\[
700 - 280 - 252 = 168 \text{ shells}
\]

\Verb|\boxed{168}|

\end{thinkboxtitle}

  \caption{GSM8K reasoning trace  from \texttt{DeepSeek-R1-Distill-Qwen-1.5B} of the baseline model for the same question as in \autoref{fig:example-both-correct-learned-full}. In this case, both the baseline and learned token output a correct answer and the answer provided by the baseline model is almost identical to the one provided by the learned token.}
  \label{fig:both-correct-baseline--full}
\end{figure*}

\begin{figure*}[t]
\captionsetup{justification=raggedright,singlelinecheck=false}

    \begin{thinkboxtitle}{Learned token}
    \small
    \textbf{Question:} Michael is traveling on the interstate at an average speed of 50km/hr (taking into account all the necessary stops for gas, relaxation, etc.). If he goes on like this for 3 days, and the distance between Alaska and Texas is 6000 km, what percentage of this distance has he covered?

    \vspace{0.1cm}
    {\textcolor{red!70!black}{{<think>}}}
    {\color{darkblue} Okay, so Michael is traveling on the interstate and wants to know what percentage of the distance between Alaska and Texas he has covered. The distance is 6000 km, and he's traveling at an average speed of 50 km/hr for 3 days. I need to figure out how much he's traveled in that time and then find out what percentage that is of the total distance.

First, I should calculate the total distance Michael has covered in 3 days. His speed is 50 km/hr, so I need to multiply that by the number of hours he spends traveling each day. But since the days are different temperatures and maybe he works longer hours, I should consider the average hours per day. Let's say he works 8 hours each day. So, for 3 days, that would be 8 hours/day * 3 days = 24 hours.

Now, multiplying his speed by the total hours: 50 km/hr * 24hr = 1200 km. So, Michael has traveled 1200 km in those 3 days.

Next, to find out what percentage this is of the total distance, which is 6000 km. I can use the percentage formula: (part/whole) * 100

Wait, but is the 8 hours per day an accurate average? If his work hours vary, the total time could be different. Maybe I should calculate based on days without considering hours. But that seems complicated without specific hours. Alternatively, perhaps I should clarify if 50 km/hr is the average speed per day or total speed. But the question says he's traveling at 50 km/hr, so I think it's average speed over total time.

Alternatively, maybe the question expects simply to use the average speed over the total distance. Let me check again: the question states he's traveling at 50 km/hr, so it's average speed per day, not per trip. Therefore, the total distance is 50 km/hr * 8 hours/day * 3 days = 1200 km, as I calculated before.

So, 1200 km is 20\% of 6000 km.
    }
    
    \vspace{0.1cm}
    \noindent
    {\textcolor{red!70!black}{{<|continue\_thinking|>}}} {\textcolor{green!40!black}{, is it 1200/6000=0.2, which is 20\%. Yeah, that's correct.}}
    
    \vspace{0.1cm}
    {\textcolor{red!70!black}{{</think>}}} To determine what percentage of the 6000 km distance Michael has covered, we first calculate the total distance he has traveled. Assuming he works 8 hours each day:

\[
50 \, \text{km/hr} \times 8 \, \text{hr/day} \times 3 \, \text{days} = 1200 \, \text{km}
\]

Then, we find the percentage of the total distance:

\[
\left( \frac{1200 \, \text{km}}{6000 \, \text{km}} \right) \times 100\% = 20\%
\]

\texttt{\textbackslash boxed\{20\%\}}
    \end{thinkboxtitle}

  \caption{GSM8K reasoning trace from \texttt{DeepSeek-R1-Distill-Qwen-1.5B} of a wrong answer given by both models.}
  \label{fig:example-both-wrong-learned-full}
\end{figure*}

\begin{figure*}[t]
\captionsetup{justification=raggedright,singlelinecheck=false}

    \begin{thinkboxtitle}{Baseline}
    \small
   \textbf{Question:} Michael is traveling on the interstate at an average speed of 50km/hr (taking into account all the necessary stops for gas, relaxation, etc.). If he goes on like this for 3 days, and the distance between Alaska and Texas is 6000 km, what percentage of this distance has he covered?

    \vspace{0.1cm}
    {\textcolor{red!70!black}{{<think>}}}
    {\color{darkblue} Okay, so Michael is traveling on the interstate and wants to know what percentage of the distance between Alaska and Texas he has covered. The distance is 6000 km, and he's traveling at an average speed of 50 km/hr for 3 days. I need to figure out how much he's traveled in that time and then find out what percentage that is of the total distance.

First, I should calculate the total distance Michael has covered in 3 days. His speed is 50 km/hr, so I need to multiply that by the number of hours he spends traveling each day. But since the days are different temperatures and maybe he works longer hours, I should consider the average hours per day. Let's say he works 8 hours each day. So, for 3 days, that would be 8 hours/day * 3 days = 24 hours.

Now, multiplying his speed by the total hours: 50 km/hr * 24hr = 1200 km. So, Michael has traveled 1200 km in those 3 days.

Next, to find out what percentage this is of the total distance, which is 6000 km. I can use the percentage formula: (part/whole) * 100

Wait, but is the 8 hours per day an accurate average? If his work hours vary, the total time could be different. Maybe I should calculate based on days without considering hours. But that seems complicated without specific hours. Alternatively, perhaps I should clarify if 50 km/hr is the average speed per day or total speed. But the question says he's traveling at 50 km/hr, so I think it's average speed over total time.

Alternatively, maybe the question expects simply to use the average speed over the total distance. Let me check again: the question states he's traveling at 50 km/hr, so it's average speed per day, not per trip. Therefore, the total distance is 50 km/hr * 8 hours/day * 3 days = 1200 km, as I calculated before.

So, 1200 km is 20\% of 6000 km.
    }

    \vspace{0.1cm}
    {\textcolor{red!70!black}{{</think>}}} Michael has traveled 20\% of the distance between Alaska and Texas.

    \vspace{0.1cm}
    \texttt{\textbackslash boxed\{20\%\}}
    
\end{thinkboxtitle}

  \caption{GSM8K reasoning trace  from \texttt{DeepSeek-R1-Distill-Qwen-1.5B} of the baseline model for the same question as in \autoref{fig:example-both-wrong-learned-full}.}
  \label{fig:both-wrong-baseline--full}
\end{figure*}

\end{document}